\documentclass[10pt,twocolumn,letterpaper]{article}

\usepackage[pagenumbers]{cvpr} 

\usepackage{bbding}
\usepackage{multirow}
\usepackage{color}
\usepackage{graphicx}
\usepackage{xcolor}
\usepackage{colortbl, booktabs}
\usepackage{xspace}
\usepackage{cuted}

\usepackage{marvosym}

\newcommand{\methodname}{Spatial-SSRL\xspace}
\usepackage{bbding}
\usepackage{multirow}
\usepackage{color}
\usepackage[most]{tcolorbox}
\usepackage{graphicx}
\usepackage{xcolor}
\usepackage{pifont}
\usepackage{ulem}
\usepackage{dsfont}
\usepackage{colortbl, booktabs}
\usepackage{comment}

\definecolor{cvprblue}{rgb}{0.21,0.49,0.74}
\usepackage[pagebackref,breaklinks,colorlinks,allcolors=cvprblue]{hyperref}

\def\paperID{5675} 
\def\confName{CVPR}
\def\confYear{2026}

\title{\methodname: Enhancing Spatial Understanding via Self-Supervised Reinforcement Learning}

\author{Yuhong Liu$^{1,2}$, Beichen Zhang$^{1,3}$, Yuhang Zang$^{1\textsuperscript{\Letter}}$, Yuhang Cao$^{1}$, Long Xing$^{1}$\\ Xiaoyi Dong$^{1}$, Haodong Duan$^{1}$, Dahua Lin$^{1, 3}$, Jiaqi Wang$^{1, 4\textsuperscript{\Letter}}$\\
$^1$Shanghai AI Laboratory \quad 
$^2$Shanghai Jiao Tong University \\
$^3$The Chinese University of Hong Kong \quad
$^4$Shanghai Innovation Institute \\
{\tt\small\{liuyuhong1,zangyuhang\}@pjlab.org.cn}
{\textsuperscript{\Letter}\small{Corresponding Authors.}}\\
{{\tt\small \textbf{Code}: \url{https://github.com/InternLM/Spatial-SSRL}}}\\
{{\tt\small \textbf{Model}: \url{https://huggingface.co/internlm/Spatial-SSRL-7B}}}\\
{{\tt\small \hspace{9.7em}\url{https://huggingface.co/internlm/Spatial-SSRL-Qwen3VL-4B}}}\\
{{\tt\small \textbf{Data}: \url{https://huggingface.co/datasets/internlm/Spatial-SSRL-81k}}}\\
}

\begin{document}
\maketitle

\begin{strip}
  \centering
  \includegraphics[width=0.98\linewidth]{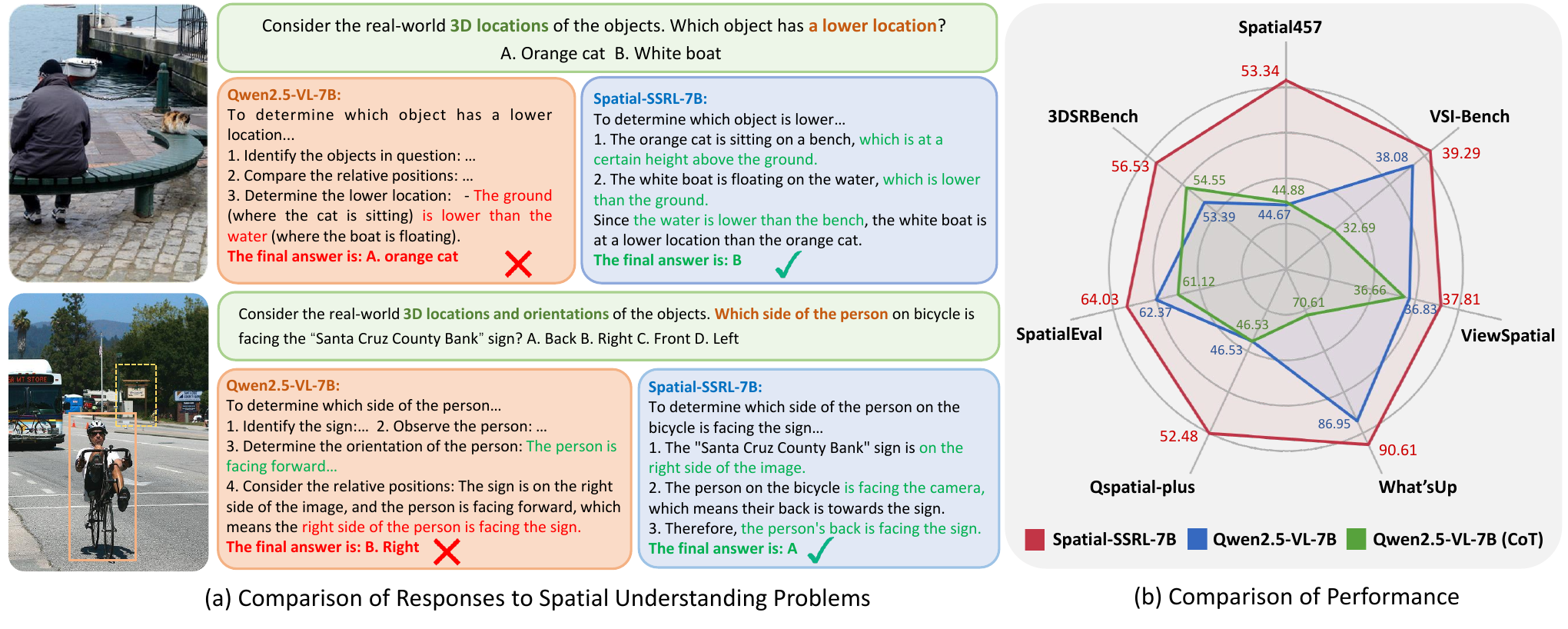}
  \vspace{-14pt}
   \captionof{figure}{We present \textbf{\methodname}, a self-supervised reinforcement learning paradigm for spatial understanding. \textbf{(a) Qualitative examples}: the baseline answers are wrong (\textcolor{red}{red}), whereas our model predicts correctly (\textcolor{ForestGreen}{green}) for 3D locations and orientations. \textbf{(b) Quantitative results} on seven spatial benchmarks show consistent improvements of \methodname-7B against Qwen2.5-VL-7B and its CoT variant.}
   \label{fig:teaser}
   \vspace{-9pt}
\end{strip}

\begin{figure*}[t]
  \centering
  \includegraphics[width=\linewidth]{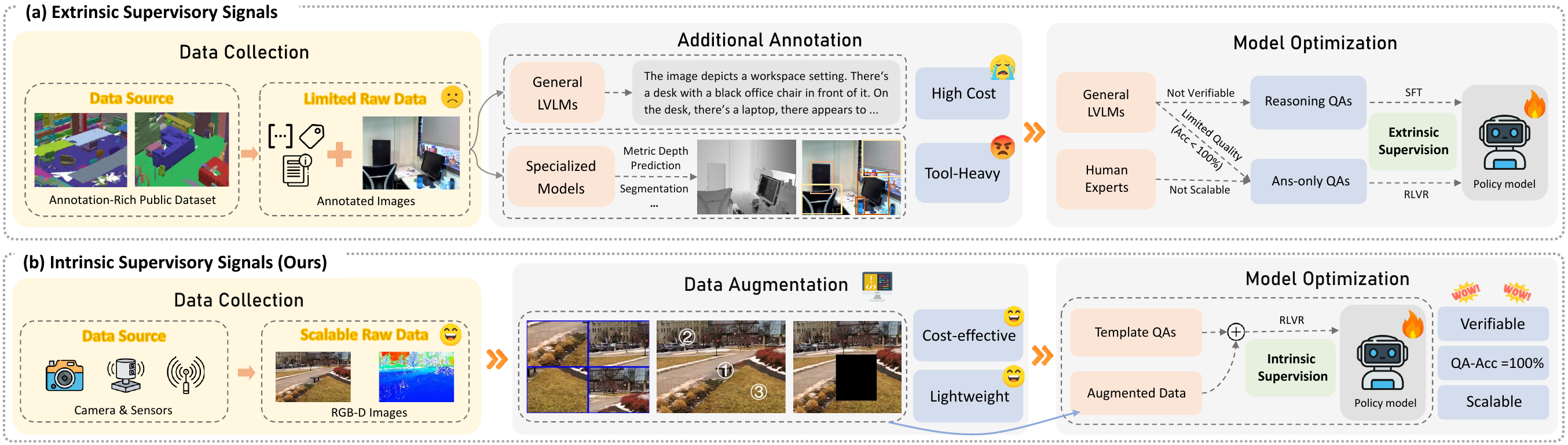}
   \vspace{-12pt}
   \caption{\textbf{(a)} Prior pipelines boost spatial understanding by injecting extrinsic supervision from expert tools or synthetic environments, which inflates cost and limits scalability. \textbf{(b)} Our \methodname replaces these dependencies with intrinsic self-supervision, yielding a scalable, lightweight, low-cost, and naturally verifiable pipeline.}
   \label{fig:comp}
   \vspace{-6pt}
\end{figure*}

\begin{abstract}
Spatial understanding remains a weakness of Large Vision-Language Models (LVLMs).
Existing supervised fine-tuning (SFT) and recent reinforcement learning with verifiable rewards (RLVR) pipelines depend on costly supervision, specialized tools, or constrained environments that limit scale.
We introduce \textbf{\methodname}, a self-supervised RL paradigm that derives verifiable signals directly from ordinary RGB or RGB-D images.
\methodname automatically formulates five pretext tasks that capture 2D and 3D spatial structure: shuffled patch reordering, flipped patch recognition, cropped patch inpainting, regional depth ordering, and relative 3D position prediction.
These tasks provide ground-truth answers that are easy to verify and require \emph{no} human or LVLM annotation.
Training on our tasks substantially improves spatial reasoning while preserving general visual capabilities.
On seven spatial understanding benchmarks in both image and video settings, \methodname delivers average accuracy gains of 4.63\% (3B) and 3.89\% (7B) over the Qwen2.5-VL baselines.
Our results show that simple, intrinsic supervision enables RLVR at scale and provides a practical route to stronger spatial intelligence in LVLMs.
\end{abstract}    
\section{Introduction}
\label{sec:intro}

Spatial understanding is pivotal for Large Vision-Language Models (LVLMs) to analyze complex real-world scenes.
As shown in \cref{fig:teaser} \textbf{(a)}, the ability to reason over \textit{depth}, \textit{distance}, \textit{azimuth}, and \textit{relative object positions} enables faithful reconstruction of 3D environments and unlocks applications such as autonomous driving \cite{tian2025nuscenes, wang2025omnidriveholisticvisionlanguagedataset}, robot manipulation \cite{kim2024openvlaopensourcevisionlanguageactionmodel, cai2025spatialbotprecisespatialunderstanding, song2025robospatialteachingspatialunderstanding, driess2023palmeembodiedmultimodallanguage}, and embodied navigation \cite{cheng2025navilaleggedrobotvisionlanguageaction, goetting2024endtoendnavigationvisionlanguage}.
Although LVLMs report near-saturated results on various tasks (e.g., visual question answering \cite{liu2024mmbench,chen2024we}, image captioning \cite{yu2024capsfusionrethinkingimagetextdata, xing2025caprlstimulatingdenseimage}, object segmentation \cite{zhang2025sec, feng2025vision}, math reasoning \cite{lu2023mathvista, zhang2025booststep}), their spatial understanding \textbf{remains substantially below} human performance \cite{li2025stibenchmllmsreadyprecise, zha2025enablellm3dcapacity, liu2025starbench, qi2025gpt4scene}.

Early data-centric Supervised Fine-Tuning (SFT) approaches \cite{chen2024spatialvlm,cheng2024spatialrgpt} advance spatial understanding by synthesizing spatial question-answer pairs from 2D images \cite{chen2024spatialvlm} or constructing single-image 3D scene graphs with a depth plugin \cite{cheng2024spatialrgpt}.
However, SFT tends to memorize dataset-specific patterns \cite{chu2025sftmemorizesrlgeneralizes,chen2025sftrlearlyinvestigation}, inherits errors from detectors and monocular depth, and often depends on expensive proprietary models for question-answering curation.
Recent Reinforcement Learning with Verifiable Reward (RLVR) methods \cite{wu2025reinforcingspatialreasoningvisionlanguage,li2025spatialladderprogressivetrainingspatial,liao2025improved} improve the generalization of spatial understanding over SFT by optimizing with verifiable rewards \cite{lambert2025tulu3pushingfrontiers, gao2024designing}, but are constrained to specific environments (e.g., 3D scans) and demand substantial pipeline engineering with limited domain coverage.
As shown in \cref{fig:comp} \textbf{(a)}, a key \textbf{open challenge} is to retain the optimization benefits of RL while scaling verifiable supervision to ordinary images across diverse domains \textbf{without} manual labels, specialized assets, or costly tooling.

To tackle this challenge, we draw on visual self-supervised learning (SSL), which learns visual representations from intrinsic structure in data via pretext objectives such as contrastive alignment and permutation/jigsaw tasks \cite{cruz2017deeppermnetvisualpermutationlearning, chen2020simpleframeworkcontrastivelearning,he2020momentum} without manual labels.
Our key idea is that intrinsic consistency signals in ordinary 2D or RGB-D images (e.g., relative depth, geometric consistency, correspondence, or invariance under view transformations) naturally supervise \textbf{spatial understanding}.
Because SSL targets are defined by the pretext itself, their correctness is deterministically verifiable during training, making SSL well aligned with the RLVR training paradigm \cite{zhang2025corewardingstableselfsupervisedrl}.
Unlike prior SSL approaches that are primarily used for visual encoder pre-training \cite{vaswani2017attention}, we repurpose SSL objectives as verifiable reward functions to directly optimize LVLM behavior via RL, shifting supervision from representation quality to spatial understanding while remaining broadly applicable across diverse image domains.

Drawing on this insight, we introduce \textbf{\methodname}, a Self-Supervised Reinforcement Learning paradigm for improving the spatial understanding of LVLMs.
According to \cref{fig:comp} \textbf{(b)}, our framework defines a suite of verifiable self-supervised tasks constructed solely from RGB-D inputs, organized into two categories: depth-free and depth-based.
\textbf{Depth-free} tasks target 2D structure, including relative position between regions, permutation ordering, and cross-view correspondence.
\textbf{Depth-based} tasks exploit depth to supervise relative depth or distance ranking and 3D relation consistency under perspective transformations.
Each task is posed as a question–answer prompt to the LVLM, with a deterministic verifier producing binary or scalar rewards.
We then optimize the model with Group Relative Policy Optimization (GRPO) \cite{shao2024deepseekmath}, using the verifiable SSL task reward and an initial cold-start stage to stabilize RL training.

Our \methodname approach is \textbf{cost-effective}, \textbf{scalable}, and \textbf{naturally compatible} with the RLVR paradigm.
Data curation is fully automated: it requires only raw RGB or RGB-D images and uses no human labels or auxiliary proprietary models.
Inputs can be collected at scale from commodity depth sensors and public sources \cite{dai2017scannet, vasiljevic2019diode, sarbolandi2015kinect}.
Because supervision comes from intrinsic image consistency rather than expensive task-specific annotations or tool chains, the pipeline generalizes across domains, is reproducible end-to-end, and is \textbf{easy to extend} with new pretext tasks without changing the data source.

As shown in \cref{fig:teaser}~\textbf{(b)}, \methodname yields substantial gains in seven spatial understanding benchmarks \cite{wang2025spatial457diagnosticbenchmark6d,ma20253dsrbenchcomprehensive3dspatial,wang2024pictureworththousandwords, liao2024reasoningpathsreferenceobjects,kamath2023whatsupvisionlanguagemodels,li2025viewspatialbenchevaluatingmultiperspectivespatial,yang2025thinkingspacemultimodallarge}.
We observe consistent accuracy improvements across benchmarks in both image and video settings.
Representative sub-tasks include relative orientation \cite{wang2025spatial457diagnosticbenchmark6d,ma20253dsrbenchcomprehensive3dspatial}, depth ordering \cite{ma20253dsrbenchcomprehensive3dspatial}, and multi-view relative location \cite{li2025viewspatialbenchevaluatingmultiperspectivespatial,kamath2023whatsupvisionlanguagemodels}.
Our largest gain is +8.67\% on Spatial457 \cite{wang2025spatial457diagnosticbenchmark6d}, which requires precise perception, strong spatial interpretation, and multi-step reasoning.
Despite the targeted RL fine-tuning, the models retain their performance on general visual question-answering (VQA), multi-image understanding, and hallucination diagnostics \cite{liu2024mmbench,fu2024blink,guan2024hallusionbench}, showing \textbf{no regression} in non-spatial capabilities.
Comprehensive ablations in \cref{sec:ablation} further indicate that both depth-free (2D layout, permutation, correspondence) and depth-based (relative depth, 3D relation) pretexts contribute to the RLVR optimization.

Our main contributions are:
\textbf{1)} We introduce \textbf{\methodname}, a self-supervised reinforcement learning paradigm for improving LVLM spatial understanding. \methodname is cost-effective, scalable, naturally compatible with the RLVR paradigm, and easy to extend to new SSL tasks.
\textbf{2)} We design a suite of self-supervised tasks for spatial perception, covering both depth-free and depth-based objectives.
Ablation studies show that each task contributes to improved spatial understanding.
\textbf{3)} We provide new insights into curating high-quality, challenging SSL data for RL training, opening a new avenue for effectively combining RLVR and SSL for improving spatial understanding.
\section{Related Work}
\label{sec:related_work}

\noindent \textbf{LVLM Spatial Understanding.}
Previous studies have been carried out to improve the spatial understanding of LVLMs. The most straightforward approach is to optimize LVLMs on high-quality question-answer (QA) pairs curated via manual annotation \cite{cai2025spatialbotprecisespatialunderstanding} or proprietary models \cite{ro2025visionlanguagemodelsunderstandcities}.
However, these approaches lack scalability due to their high cost.
A more affordable alternative uses public datasets (e.g., ScanNet \cite{dai2017scannet}) with abundant meta-information \cite{ouyang2025spacerreinforcingmllmsvideo}, but the scale and richness of the curated data are inherently constrained by the datasets employed.

To construct large-scale spatial QAs in a cost-effective manner, two dominant paradigms have emerged: tool-based and simulation-based approaches.
Tool-based approaches incorporate tools within their pipelines. The tools are either open-source models that struggle with spatial understanding \cite{cheng2024spatialrgpt}, or expert models \cite{chen2024spatialvlm,hong20233dconceptlearningreasoning,deng2025internspatialcomprehensivedatasetspatial} for object detection, segmentation, and depth estimation, resulting in an overly complex pipeline with additional computational cost (depicted in \cref{fig:comp}). Simulation-based approaches render 3D scenes and synthesize QAs \cite{zhang2025mllmsstrugglespatialunderstanding, zhang2025spatialunderstandingvideosstructured}, whose quality remains unsatisfactory due to divergence from real-world scenarios. In contrast, \methodname provides a novel paradigm featuring a tool-free pipeline, real-world consistency, cost-effectiveness, and high scalability.

\noindent \textbf{Self-Supervised Learning.}
Self-supervised learning (SSL), which obtains supervision from the inherent structure of the data itself without relying on manual labels \cite{gui2024surveyselfsupervisedlearningalgorithms}, has been remarkably effective for visual representation learning. Early approaches learn visual features via contrastive learning \cite{chen2020simpleframeworkcontrastivelearning, he2020momentum} or self-supervised tasks such as rotation \cite{gidaris2018unsupervisedrepresentationlearningpredicting}, jigsaw \cite{noroozi2017unsupervisedlearningvisualrepresentations}, and temporal ordering \cite{misra2016shufflelearnunsupervisedlearning}. With the rise of LLMs and LVLMs, SSL has become prevalent in pre-training.
Pre-training of autoregressive LMs (GPT \cite{brown2020languagemodelsfewshotlearners}), masked language models (BERT \cite{devlin2019bertpretrainingdeepbidirectional}), and masked autoencoders (MAE \cite{he2021maskedautoencodersscalablevision}) all follow the paradigm of adding masks to parts of the data and learning to predict the masked content. However, strategies for using SSL in the LVLM post-training phase remain limited.

Recently, several approaches appear to enhance LVLMs via self-supervised post-training \cite{wu2025visualjigsawposttrainingimproves, guo2025ssl4rlrevisitingselfsupervisedlearning, wang2025jigsaw}. Jigsaw-R1 \cite{wang2025jigsaw} uses jigsaw puzzles for RL. Visual Jigsaw \cite{wu2025visualjigsawposttrainingimproves} is a concurrent work that constructs jigsaw tasks across three different modalities. Our work covers a broader set of self-supervised tasks, with Jigsaw as only one component.
SSL4RL \cite{guo2025ssl4rlrevisitingselfsupervisedlearning} focuses solely on 2D tasks, while our work targets spatial understanding and designs verifiable supervisory signals based on both 2D and RGB-D images.

\noindent \textbf{Reinforcement Learning with Verifiable Rewards.}
Following the success of Deepseek-R1 \cite{guo2025deepseek}, many recent works have applied Group Relative Policy Optimization \cite{shao2024deepseekmath} and demonstrated the potential of reinforcement learning with verifiable rewards (RLVR) \cite{ouyang2025spacerreinforcingmllmsvideo, wang2025reinforcement, liu2025visual, wen2025reinforcement, zhang2025right}. For example, Visual-RFT improves performance on image classification, detection, and grounding \cite{liu2025visual}, while other approaches incentivize math reasoning via RLVR \cite{zhang2025right,wang2025reinforcement}.

In the realm of spatial understanding, existing methods mainly rely on annotation-rich public datasets and tool-heavy pipelines \cite{liao2025improved,li2025spatialladderprogressivetrainingspatial,ouyang2025spacerreinforcingmllmsvideo} to curate training data for RL. Such approaches are unable to fully exploit the advantages of RL because the quality and scale of their data are limited by the specialized assets they employ. To address this problem, we introduce a novel paradigm that scales verifiable supervision and enhances LVLM spatial understanding.
\section{\methodname} 
\begin{figure*}[t]
  \centering
  \includegraphics[width=\linewidth]{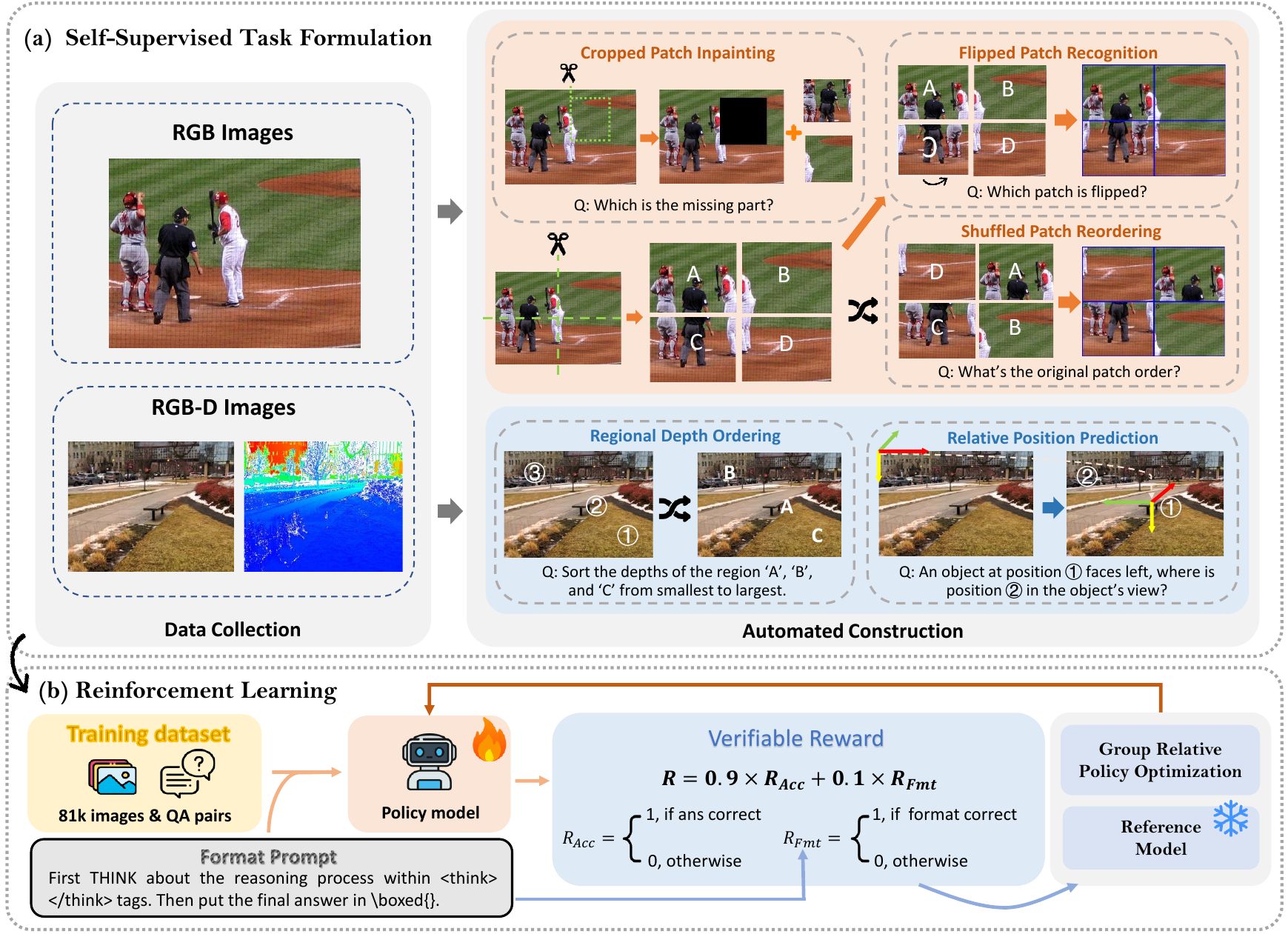}
   \caption{\textbf{Overview of \methodname.} \textbf{(a) Self-supervised data curation}: from raw RGB and RGB-D images, we automatically construct five pretext tasks (patch reordering, patch flip detection, cropped-patch inpainting, regional depth ordering, and relative 3D position prediction), requiring no human or LLM annotations. \textbf{(b) RL training}: the model is optimized with Group Relative Policy Optimization (GRPO) using a verifiable reward function that evaluates answer correctness, and a format reward that elicits format compliance.}
   \vspace{-12pt}
   \label{fig:overview}
   \vspace{-6pt}
\end{figure*}

The overview of \methodname is shown in \cref{fig:overview}.
Our framework consists of two stages: self-supervised task design (\cref{section:self-supervised-tasks}) and reinforcement learning (\cref{section:training}).
The design is guided by three principles that distinguish our approach from prior work: (i) \textit{zero human or LLM supervision}: all ground-truth labels are derived deterministically from image structure; (ii) \textit{tool-free scalability}: no external detection, segmentation, or rendering pipelines are required; and (iii) \textit{natural verifiability}: supervision signals are verifiable and suitable for RL reward computation.

\noindent \textbf{Task Design.}
To comprehensively enhance spatial understanding, we design five complementary pretext tasks in total.
\textit{Shuffled patch reordering} requires recovering global 2D layout consistency from permuted patches.
\textit{Flipped patch recognition} demands sensitivity to mirror symmetries and local orientation cues.
\textit{Cropped patch inpainting} tests texture-context matching and fine-grained structural reasoning.
\textit{Regional depth ordering} evaluates ordinal depth perception across image regions.
\textit{Relative 3D position prediction} assesses egocentric spatial relations (left/right, front/back) conditioned on object orientation.
These tasks jointly encourage both 2D layout understanding and 3D spatial reasoning, providing unambiguous targets for RLVR optimization.
Importantly, the \methodname framework is \textit{modular} and \textit{extensible}: additional self-supervised tasks can be integrated without modifying the RL pipeline, enabling future work to expand spatial coverage.

\subsection{Self-Supervised Task Design}
\label{section:self-supervised-tasks}
We organize our five self-supervised tasks into two complementary categories that target different aspects of spatial intelligence.
The three \textbf{depth-free} tasks (\cref{section:depth-free}) operate solely on RGB images and emphasize 2D spatial relationships, including patch-level layout, structural consistency, and fine-grained texture-context correspondence.
The two \textbf{depth-based} tasks (\cref{section:depth-based}) use depth information to supervise 3D scene understanding, focusing on ordinal depth perception and egocentric spatial relations.
Together, these tasks provide comprehensive coverage of spatial reasoning capabilities.

\subsubsection{Depth-free Tasks}
\label{section:depth-free}
RGB images contain rich intrinsic structural information that can serve as self-supervision without external annotation.
By applying deterministic augmentation operations solely to pixel-level content, we construct Question-Answer (QA) pairs for three complementary tasks: \textit{Shuffled patch reordering}, \textit{Flipped patch recognition}, and \textit{Cropped patch inpainting}.
Each task exploits different aspects of image structure to promote spatial understanding.

\noindent \textbf{Shuffled Patch Reordering.} 
\textit{Motivation}: Recovering the original order of shuffled patches requires understanding global 2D layout consistency and relative positional relationships, skills that directly transfer to reasoning about object arrangements in real scenes.

\textit{Formulation}: Given an image $\mathit{I}\in \mathbb{R}^{H\times W}$, we partition it into $M\times N$ grid of patches, each of size $P_H\times P_W$, where $P_H=\frac{H}{M}$ and $P_W=\frac{W}{N}$.
We denote the patch grid as $\mathcal{X}=\{x_{i,j} \in \mathbb{R}^{P_H \times P_W} \mid i\in[0, M), j\in[0, N)\}$, where $x_{i,j}=\mathit{I}(iP_H:(i+1)P_H, jP_W:(j+1)P_W)$.
For convenience, we flatten $\mathcal{X}$ into a sequence $\hat{\mathcal{X}}=[\hat{x}_0, \hat{x}_1, \ldots, \hat{x}_{M\times N-1}]$ with $\hat{x}_k=x_{i,j}$ and $k=iN+j$.
We then apply a random permutation $\pi:\{0, \ldots, M\times N-1\}\mapsto \{0, \ldots, M\times N-1\}$ to generate the shuffled sequence $\hat{\mathcal{X}}_{\text{shuffle}}=[\hat{x}_{\pi(0)}, \hat{x}_{\pi(1)}, \ldots, \hat{x}_{\pi(M\times N-1)}]$, from which we reconstruct the shuffled input image $I_{\text{input}}$.

\textit{Task objective}: The model is prompted to predict the correct patch ordering that restores the original image.
Since $\pi$ is bijective, its inverse $\pi^{-1}$ exists and provides the ground-truth answer:
\begin{equation}
    \pi^{-1}=[\pi^{-1}(0), \pi^{-1}(1), \ldots, \pi^{-1}(M\times N-1)].
\end{equation}

\textit{Difficulty enhancement}: To increase task complexity and prevent trivial edge-matching solutions, we optionally mask one random patch $\hat{x}_{\pi(t)}$ by setting all pixels to white: $\hat{x}_{\pi(t)}(0:P_{H}, 0:P_{W})=255$.
This forces the model to rely on global layout rather than local patch boundaries.
Implementation details are provided in Appendix \ref{sec:app-shuf}.

\noindent \textbf{Flipped Patch Recognition.}
\textit{Motivation}: Detecting subtle orientation violations requires sensitivity to local geometry, mirror symmetries, and directional cues such as text, faces, and shadows.
These capabilities are essential for understanding viewpoint-dependent spatial relations.

\textit{Formulation}: Given an image $I\in \mathbb{R}^{H\times W}$, we partition it into patches using the same grid structure as shuffled patch reordering, yielding $\hat{\mathcal{X}}=\{\hat{x}_k\}$ with $k\in[0, M\times N-1]$.

We randomly select one patch $\hat{x}_t$ and apply a flip operation $f: \hat{x}_t \mapsto \hat{x}_t^{\text{flip}}$ defined as:
\begin{equation}
\begin{cases}
x_{\text{vert}}, & \text{with probability } 0.5, \\
x_{\text{horz}}, & \text{with probability } 0.5,
\end{cases}
\end{equation}
where $x_{\text{vert}}(r, c)=x(P_H-1-r, c)$ (vertical flip) and $x_{\text{horz}}(r, c)=x(r, P_W-1-c)$ (horizontal flip).
The input image $I_{\text{input}}$ is reconstructed from the modified sequence $\hat{\mathcal{X}}'=[\hat{x}_0, \ldots, \hat{x}_t^{\text{flip}}, \ldots, \hat{x}_{M\times N-1}]$.

\textit{Task objective}: The model must identify both the index $t$ of the flipped patch and its flip direction $d\in\{\text{vert}, \text{horz}\}$.
The ground-truth answer is the tuple $[t, d]$.

\noindent
\textbf{Cropped Patch Inpainting}.
\textit{Motivation}: Identifying which patch correctly fills a masked region requires analyzing texture continuity, semantic context, and structural consistency between local content and its surroundings.
These skills generalize to understanding spatial coherence.

\textit{Formulation}: Given an image $I\in \mathbb{R}^{H\times W}$, we extract a square patch of side length $s=\min(H/2, W/2)$ whose top-left corner $(x_0, y_0)$ is sampled uniformly:
\begin{equation}
    (x_0, y_0)\sim \mathcal{U}\big([0, H-s]\times[0, W-s]\big).
\end{equation}
The cropped region is $\mathcal{R}=[x_0, x_0+s]\times[y_0, y_0+s]$, and we denote the extracted patch as $I_{\text{crop}}=I(\mathcal{R})$.
We construct the masked input image $I_{\text{input}}$ by zeroing out the cropped region:
\begin{equation}
    I_{\text{input}}(u, v) =
    \begin{cases}
    0, & (u, v)\in \mathcal{R}, \\
    I(u, v), & \text{otherwise}.
    \end{cases}
\end{equation}

\textit{Task objective}: The model is presented with $I_{\text{input}}$ and four candidate patches in a multiple-choice format, and must select the correct patch $I_{\text{crop}}$ that fills the masked region.

\textit{Distractor construction}: To prevent trivial solutions based on low-level texture matching, we construct three challenging distractors that share substantial visual similarity with the ground-truth patch.
The three distractors are: (1) a $90^\circ$ rotated version of $I_{\text{crop}}$; (2) the interior subregion $I_{\text{int}}=I_{\text{crop}}(\mathcal{R}_{\text{int}})$, where $\mathcal{R}_{\text{int}}=[\frac{s}{4}, \frac{3s}{4}-1]\times[\frac{s}{4}, \frac{3s}{4}-1]$; and (3) the exterior region $I_{\text{ext}}=I(\mathcal{R}_{\text{ext}})$, where $\mathcal{R}_{\text{ext}}=[x_0-\theta s, x_0+(1+\theta)s]\times[y_0-\theta s, y_0+(1+\theta)s]$ with $\theta\in\{0.25, 0.5\}$.
For $\mathcal{R}_{\text{ext}}$ extending beyond image boundaries, we clip to $[0, H]\times[0, W]$.
All distractors are resized to $s\times s$ to prevent size-based discrimination, forcing the model to attend to fine-grained structural and semantic consistency.
Please refer to Appendix~\ref{sec:app-crop} for more details.

\subsubsection{Depth-based Tasks}
\label{section:depth-based}
Depth maps encode per-pixel distances from the camera, providing rich geometric supervision for 3D spatial understanding without requiring manual annotation.
We design two complementary depth-based tasks that supervise spatial reasoning from different perspectives: \textit{Regional depth ordering} adopts a camera-centric view and tests ordinal depth perception across regions, while \textit{Relative position prediction} adopts an object-centric view and evaluates egocentric spatial relations conditioned on object orientation.
Together, these two tasks promote robust 3D scene comprehension.

\noindent
\textbf{Regional Depth Ordering.}
\textit{Motivation}: Ranking regions by distance from the camera requires integrating depth cues, perspective understanding, and ordinal reasoning, which are foundational skills for tasks such as occlusion reasoning and 3D scene reconstruction.

\textit{Formulation}: Given an image $I\in \mathbb{R}^{H\times W}$ and its normalized depth map $D\in \mathbb{R}^{H\times W}$, we select three disjoint regions $R_1, R_2, R_3 \subseteq [0, H]\times[0, W]$ with increasing depth (i.e., $R_1$ is closest to the camera, $R_3$ is farthest).
To ensure unambiguous ordering, we enforce the following constraints:
\begin{equation}
    r(R_i)=\max\limits_{(x, y)\in R_i}D(x,y)-\min\limits_{(x, y)\in R_i}D(x,y)<r_{max},
\end{equation}
\begin{equation}
    d(R_i, R_{i+1})=\min\limits_{(x, y)\in R_{i+1}}D(x,y)-\max\limits_{(x, y)\in R_i}D(x,y)>d_{min},
\end{equation}
where $r(R_i)$ denotes the depth range within region $R_i$ and $d(R_i, R_{i+1})$ denotes the depth gap between consecutive regions.
We set $r_{\max}=0.15$ and $d_{\min}=0.05$ to guarantee well-separated regions with consistent internal depth.
We then apply a random permutation $\hat{\pi}:\{1,2,3\}\mapsto\{1,2,3\}$ to assign visual labels, marking each region's center with label $\hat{\pi}(R_i)$ on the image to construct $I_{\text{input}}$.

\textit{Task objective}: The model must order the labeled regions from closest to farthest.
The ground-truth answer is the sequence $\hat{\pi}=[\hat{\pi}(1), \hat{\pi}(2), \hat{\pi}(3)]$.

\noindent \textbf{Relative Position Prediction.}
\textit{Motivation}: Determining spatial relations from an object's perspective (e.g., ``the cup is to the left of the book'') requires mental rotation, egocentric coordinate transformation, and integration of orientation cues with depth information.

\textit{Formulation}: Given an image with depth map $D$, we select two pixel locations $R_1$ and $R_2$.
We assume an object is located at $R_1$ with a specified orientation and the model is required to predict the relative position of $R_2$ from the object's viewpoint (e.g., \textit{left}, \textit{right-front}).
The task is posed as a multiple-choice question with four options.
The construction process and coordinate definitions are shown in \cref{fig:orientation}.

\textit{Coordinate transformation}: Let $(x_1, y_1, z_1)$ and $(x_2, y_2, z_2)$ denote the camera-frame coordinates of $R_1$ and $R_2$, where $z_i=D(x_i, y_i)$.
The object's orientation is defined by a rotation angle $\theta$ (counterclockwise from the camera's $z$-axis to the object's forward direction).
We assume the object's orientation is parallel to the ground plane, effectively projecting the problem onto the $xz$-plane and ignoring vertical displacement, as $y$-coordinates do not reliably encode real-world height due to perspective projection.

The orientation $\theta$ is sampled uniformly from \{\textit{left} ($90^\circ$), \textit{towards camera} ($180^\circ$), \textit{right} ($270^\circ$), \textit{away from camera} ($0^\circ$)\}.
To compute the relative position of $R_2$ in the object's coordinate frame, we apply a 2D rigid transformation (translation followed by rotation) to $(x_2, z_2)$:
\begin{small}
\begin{equation}
\left[
\begin{array}{c}
   \tilde{x}_2 \\
   \tilde{z}_2 \\
   1
\end{array}
\right]
=
\underbrace{\left[
\begin{array}{ccc}
    \cos\theta  &\sin\theta & 0 \\
    -\sin\theta& \cos\theta & 0 \\
    0 & 0 & 1
\end{array}
\right]}_{\text{Rotation}}
\underbrace{
\left[
\begin{array}{ccc}
    1  & 0 & -x_1 \\
    0 & 1 & -z_1 \\
    0 & 0 & 1
\end{array}
\right]}_{\text{Translation}}
\left[
\begin{array}{c}
    x_2 \\
    z_2 \\
    1
\end{array}
\right]. 
\end{equation}
\end{small}

\textit{Task objective}: The relative position is determined by the signs of $(\tilde{x}_2, \tilde{z}_2)$ in the object's coordinate frame.
We define the directional labels as:
\begin{equation}
\tilde{p}_x =
\begin{cases}
\text{Right}, & \tilde{x}_2>\delta_x\\
\text{Left}, & \tilde{x}_2<-\delta_x \\
\text{None}, & \text{otherwise}
\end{cases}, \quad
\end{equation}
\begin{equation}
\tilde{p}_z =
\begin{cases}
\text{Front}, & \tilde{z}_2>\delta_z\\
\text{Back}, & \tilde{z}_2<-\delta_z \\
\text{None}, & \text{otherwise}
\end{cases},
\end{equation}
where $\delta_x$ and $\delta_z$ are thresholds that enforce unambiguous spatial separation.
We discard instances where both $\tilde{p}_x=\tilde{p}_z=\text{None}$, ensuring all valid answers describe a clear spatial relation along at least one axis.
The ground-truth answer is the tuple $(\tilde{p}_x, \tilde{p}_z)$.
Additional details are provided in Appendix \ref{sec:app-pos}.

\subsubsection{Dataset Construction}\label{subsec:dataset}
\noindent \textbf{Data sources.} We collect raw RGB images from COCO~\cite{lin2014microsoft} and RGB-D images from DIODE~\cite{vasiljevic2019diode} and MegaDepth~\cite{li2018megadepth}.
These datasets provide real-world imagery spanning diverse indoor and outdoor scenes, object categories, and viewpoints.
Critically, we use \textit{only} the raw images and depth maps (where available), discarding all human-provided annotations such as bounding boxes, segmentations, or captions.
This ensures our pipeline remains fully self-supervised and reproducible without dependence on costly annotation infrastructure.

\noindent \textbf{\methodname-81k dataset.} By applying the automated augmentation and task construction procedures described in \cref{section:self-supervised-tasks}, we generate a dataset of 81,053 question-answer pairs, denoted \textbf{\methodname-81k}.
The dataset balances depth-free and depth-based tasks in roughly equal proportions and exhibits diverse question formats, including ordering tasks, multiple-choice questions with image options, and multiple-choice questions with text options.
Importantly, because all supervision is derived deterministically from image structure, \methodname-81k achieves 100\% ground-truth accuracy, which is unattainable by prior pipelines that rely on noisy detections or model-generated annotations.
Detailed statistics, task distributions, and example QA pairs are provided in Appendix \ref{sec:app-81k}.

\subsection{RL Training with Verifiable Rewards}
\label{section:training}

We optimize LVLMs on the \methodname-81k dataset using Group Relative Policy Optimization (GRPO) \cite{shao2024deepseekmath}, a policy-gradient RL algorithm well-suited for verifiable reward signals.
Our training procedure has two stages: an SFT cold-start phase followed by GRPO optimization.

\noindent \textbf{Cold-start with SFT.}
Our five self-supervised tasks vary substantially in difficulty and output format (ordering sequences, multiple-choice with image or text options).
In preliminary experiments, we observe that directly applying RL from a pretrained checkpoint leads to training instability and reward collapse, as the base model fails to generate valid formatted responses (success rate $<5\%$ for complex tasks like relative position prediction).
To mitigate this, we first perform a brief SFT warm-up on a small subset of approximately 3,600 samples ($\sim$4.4\% of the full dataset).
This cold-start phase familiarizes the model with task formats and answer structures while preserving the benefits of RL-based optimization in the subsequent stage.

\noindent \textbf{GRPO Optimization.}
Following cold-start, we apply GRPO to optimize the policy on all five tasks jointly.
We append a format prompt to each question, instructing the model to generate its reasoning process within $\langle$\texttt{think}$\rangle$ $\ldots$ $\langle$/\texttt{think}$\rangle$ tags and provide the final answer in \textbackslash\texttt{boxed}\{\}.
Our structured output format encourages step-by-step reasoning and enables reliable answer extraction for reward computation.

\noindent \textbf{Reward Design.}
The GRPO reward function has two binary components: an accuracy reward $r_{\text{acc}}$ and a format reward $r_{\text{fmt}}$.
The \textit{accuracy reward} is set to $r_{\text{acc}}=1$ if and only if the model's predicted answer exactly matches the ground-truth (determined via self-supervision), and $r_{\text{acc}}=0$ otherwise.
The \textit{format reward} is $r_{\text{fmt}}=1$ if the output strictly adheres to the prescribed format (reasoning enclosed in think tags, answer in boxed command), and $r_{\text{fmt}}=0$ otherwise.
The overall reward is a weighted combination, $r = 0.9 \cdot r_{\text{acc}} + 0.1 \cdot r_{\text{fmt}}.$
We assign higher weight to accuracy than format because format compliance typically stabilizes quickly after cold-start ($>95\%$ compliance), whereas accuracy improvements require the full RL phase.
\section{Experiments}
\label{sec:exp}

\begin{table*}[t]
  \caption{\textbf{Performance of Qwen2.5-VL-based models on spatial understanding benchmarks.} The open-source models and our model are evaluated on seven benchmarks, and the average results are provided in the last column. For a comprehensive comparison, we test two settings of the baseline model: one that does not generate the reasoning process and one that does. We compute the improvements based on the results of our models and the baseline models without reasoning. The qualitative analysis of some cases is shown in Appendix~\ref{sec:app-qualitative}.}
  \centering
  \vspace{-10pt}
  \setlength{\tabcolsep}{3pt}
  \scalebox{0.95}{
  \normalsize
  \begin{tabular}{@{}lccccccccc@{}}
    \toprule
    \multirow{2}{*}{Models} & \multirow{2}{*}{Reasoning} & \multicolumn{6}{c}{Image} & \multicolumn{1}{c}{Video}  & \multirow{2}{*}{Avg.}  \\
\cmidrule(lr){3-8} \cmidrule(l){9-9} 
  & & Spatial457 & 3DSRBench & SpatialEval & $\text{QSpatial}_{plus}$ &{What'sUp}& ViewSpatial  & {VSI-Bench}\\
    \midrule
    \rowcolor{blue!8}
    \multicolumn{10}{c}
    {\textit{Representative Spatial Models}} \\
    \cmidrule(){1-10}
    SpatialLadder-3B \cite{li2025spatialladderprogressivetrainingspatial} & \Checkmark & 40.19 &48.91 &41.57 &37.62& 66.59 &39.06 & 41.67& 45.09 \\
    SpaceR-7B \cite{ouyang2025spacerreinforcingmllmsvideo} & \Checkmark & 53.66 & 53.91 & 61.57 & 48.51 & 86.22 &  37.34 & 40.54 & 54.54 \\
    \midrule
    \rowcolor{blue!8}
    \multicolumn{10}{c}{\textit{Baselines \&  Our Model (3B)}} \\
    \cmidrule(){1-10}
    Qwen2.5-VL-3B & \XSolidBrush & 33.70 & 50.30 & 54.65 & 33.66 & 85.85 & 35.38  & 27.84 & 45.91  \\
    Qwen2.5-VL-3B & \Checkmark & 34.82 & 49.47 & 50.55 & 30.66 & 83.54 & 34.23 & 30.67 & 44.85\\
    \methodname-3B & \Checkmark & 46.07 & 51.72 & 59.59 & 39.60 & 86.71 & 36.62 & 33.49 & 50.54 \\
    \rowcolor{gray!10}
    \textit{Improvement} & \textit{N/A} & 
    \textcolor[RGB]{0, 150, 50}{+12.37} &
    \textcolor[RGB]{0, 150, 50}{+1.42} &
    \textcolor[RGB]{0, 150, 50}{+4.94} &
    \textcolor[RGB]{0, 150, 50}{+5.95} &
    \textcolor[RGB]{0, 150, 50}{+0.86} &
    \textcolor[RGB]{0, 150, 50}{+1.24}&
    \textcolor[RGB]{0, 150, 50}{+5.65} &
    \textbf{\textcolor[RGB]{0, 120, 50}{+4.63}} \\
    \midrule 
    \rowcolor{blue!8}
    \multicolumn{10}{c}{\textit{Baselines \& Our Model (7B)}} \\
    \cmidrule(){1-10}
    Qwen2.5-VL-7B & \XSolidBrush & 44.67 & 53.39 & 62.37 & 46.53 & 86.95 & 36.83  & 38.08 & 52.69  \\
    Qwen2.5-VL-7B & \Checkmark & 44.88 & 54.55 & 61.12 & 46.53 & 70.61 & 36.66 & 32.69& 49.58\\
    \methodname-7B & \Checkmark & 53.34 & 56.53 & 64.03 & 54.46 & 90.61& 37.81 & 39.29 & 56.58 \\
    \rowcolor{gray!10}
    \textit{Improvement} & \textit{N/A} & 
    \textcolor[RGB]{0, 150, 50}{+8.67} &
    \textcolor[RGB]{0, 150, 50}{+3.14} &
    \textcolor[RGB]{0, 150, 50}{+1.66} &
    \textcolor[RGB]{0, 150, 50}{+7.93} &
    \textcolor[RGB]{0, 150, 50}{+3.66} &
    \textcolor[RGB]{0, 150, 50}{+0.98}&
    \textcolor[RGB]{0, 150, 50}{+1.21} &
    \textbf{\textcolor[RGB]{0, 120, 50}{+3.89}} \\
    \bottomrule
  \end{tabular}
  }
  \vspace{-8pt}
  \label{tab:spa}
\end{table*}

\begin{table}[h]
  \caption{\textbf{Performance of Qwen-3-VL-based models on spatial understanding and general VQA benchmarks.} We evaluate the models on the seven benchmarks in \cref{tab:spa} and the four \textit{General VQA} benchmarks in \cref{tab:general}, and show the average results. The detailed results are shown in Appendix~\ref{sec:app-qwen3}. }
  \centering
  \vspace{-10pt}
  \setlength{\tabcolsep}{3pt}
  \scalebox{0.98}{
  \normalsize
  \begin{tabular}{@{}l|c|c|c|c@{}}
    \toprule
    Models & CoT & Spatial Und. & CoT & Gnr.-VQA\\
    \midrule
    Qwen3-VL-4B & \XSolidBrush & 60.14 & \XSolidBrush & 69.10 \\
    Qwen3-VL-4B & \Checkmark & 60.23 & \textit{N/A} & \textit{N/A} \\
    {\methodname}-4B & \Checkmark & \textbf{61.43} \textbf{\textcolor[RGB]{0, 120, 50}{(+1.29)}} & \XSolidBrush & \textbf{70.28} \textbf{\textcolor[RGB]{0, 120, 50}{(+1.18)}}  \\
    \bottomrule
  \end{tabular}
  }
  \vspace{-12pt}
  \label{tab:spa_v3}
\end{table}

\subsection{Experimental Settings}\label{sec:exp_settings}
\textbf{Base models and training.}
We train \textbf{\methodname-3B} and \textbf{\methodname-7B} from Qwen2.5-VL-3B and Qwen2.5-VL-7B \cite{bai2025qwen2}, and train \textbf{\methodname-4B} from Qwen3-VL-4B on our \methodname-81k.
In the cold-start stage, we train for 5 epochs on the SFT data with a learning rate of $1\times10^{-5}$.
During GRPO, we apply KL regularization with weight 0.01.
For each training sample, we generate a rollout group of 5 responses with temperature 1.0.
The training uses a global batch size of 128 and a learning rate of $1\times 10^{-6}$ for 360 steps.

\noindent
\textbf{Evaluation Protocol.}
We evaluate all models using VLMEvalKit \cite{duan2025vlmevalkitopensourcetoolkitevaluating}, an open-source toolkit that provides standardized evaluation protocols and metrics for vision-language models.
For benchmarks not natively supported by VLMEvalKit, we adapt the official evaluation code to accommodate the Qwen2.5-VL architecture while strictly preserving the original metrics and evaluation procedures.
When evaluating models with reasoning capabilities (\methodname-3B and \methodname-7B), we append the same format prompt used during GRPO training (\cref{section:training}) to elicit structured reasoning, ensuring consistency between training and inference.
For fair comparison, baseline models without reasoning are evaluated using their standard prompts without modification.
Additional evaluation details, including prompt templates and per-benchmark configurations, are provided in Appendix \ref{sec:app-exp}.

\subsection{Main Results}
\subsubsection{Spatial Understanding}
\label{sec:exp-spa}
\noindent \textbf{Benchmark overview.}
We evaluate spatial understanding across seven diverse benchmarks spanning image and video modalities.
Image benchmarks include Spatial457 \cite{wang2025spatial457diagnosticbenchmark6d}, 3DSRBench \cite{ma20253dsrbenchcomprehensive3dspatial}, SpatialEval \cite{wang2024pictureworththousandwords}, QSpatial-plus \cite{liao2024reasoningpathsreferenceobjects}, What'sUp \cite{kamath2023whatsupvisionlanguagemodels}, and ViewSpatial \cite{li2025viewspatialbenchevaluatingmultiperspectivespatial}, while VSI-Bench \cite{yang2025thinkingspacemultimodallarge} focuses on understanding egocentric videos. The tasks in these benchmarks cover a variety of types (distance estimation, object occlusion, collision prediction, object orientation, multi-view understanding, etc.) and require strong 2D and 3D visual-spatial intelligence of LVLMs.

\noindent \textbf{Overall performance improvements.} As shown in \cref{tab:spa} and \cref{tab:spa_v3}, \methodname improves spatial understanding on both Qwen2.5-VL and Qwen3-VL architectures. Importantly, \methodname achieves consistent improvements on Qwen2.5-VL across all seven benchmarks for both 3B and 7B model scales.
The 3B model yields an average gain of \textbf{+4.63\%}, with particularly strong improvements on Spatial457 (+12.37\%), and VSI-Bench (+5.65\%).
The 7B model also achieves an average gain of \textbf{+3.89\%}.
The substantial gains on Spatial457, which requires complex pose estimation and multi-step spatial reasoning, demonstrate that \methodname enhances not only spatial perception but also higher-level spatial reasoning capabilities.

\noindent \textbf{Reasoning capability enhancement.}
A critical observation from \cref{tab:spa} is that the baseline models exhibit performance degradation when prompted to generate explicit reasoning chains.
Specifically, Qwen2.5-VL-3B drops from 45.91\% to 44.85\% average accuracy with reasoning enabled, while Qwen2.5-VL-7B shows mixed results with notable degradation on What'sUp (86.95\% $\rightarrow$ 70.61\%) and VSI-Bench (38.08\% $\rightarrow$ 32.69\%).
This phenomenon, consistent with prior work~\cite{wu2025reinforcingspatialreasoningvisionlanguage, zhang2025mllmsstrugglespatialunderstanding}, indicates that base models lack effective spatial reasoning and their generated reasoning steps introduce noise rather than providing useful inference.
In contrast, \methodname models with reasoning consistently outperform baselines without reasoning across all benchmarks, demonstrating that our self-supervised RL paradigm successfully teaches the model to generate productive spatial reasoning rather than spurious correlations.

\noindent \textbf{Analysis of improvement patterns.}
The improvement distribution across benchmarks reveals meaningful patterns.
The largest gains appear on benchmarks requiring 3D spatial reasoning (Spatial457, 3DSRBench), validating that both depth-free and depth-based pretext tasks contribute to 3D understanding.
Moderate gains on 2D-centric benchmarks (SpatialEval, What'sUp) suggest that depth-free tasks like patch reordering and inpainting effectively enhance 2D spatial layout comprehension.
The relatively smaller but consistent improvements on ViewSpatial indicate that orientation and multi-perspective understanding benefit from our pretext tasks as well.

\noindent \textbf{Cross-modal generalization.} Despite training exclusively on static RGB and RGB-D images, \methodname improves video spatial understanding on VSI-Bench by +5.65\% (3B) and +1.21\% (7B).
The ability to transfer from image-based self-supervision to video understanding suggests that the pretext tasks foster internal spatial representations that are modality-agnostic and grounded in geometric principles.

\begin{table*}[t]
\caption{\textbf{General visual capability comparisons.} The benchmarks are organized into two categories: \textit{General VQA} and \textit{OCR and Chart Understanding}. The first category covers a wide range of fundamental visual capabilities, such as knowledge application, and hallucination recognition, multi-image understanding, etc. The second category targets the understanding of images with charts or rich textual details. The average accuracy of both categories are computed and provided.}
  \centering
  \vspace{-10pt}
  \setlength{\tabcolsep}{3pt}
  \scalebox{0.84}{
  \begin{tabular}{@{}lccccccccc@{}}
    \toprule
    \multirow{2}{*}{Models}  & \multicolumn{5}{c}{General VQA} & \multicolumn{4}{c}{OCR and Chart Understanding}   \\
    \cmidrule(lr){2-6} \cmidrule(l){7-10}  
    & MMBench1.1 & BLINK & Hallusion & RealWorldQA & Avg. &
  OCRBench &ChartQA&
  $\text{SeedBench2}_{plus}$  & Avg.\\
    \cmidrule(){1-10} 
    \rowcolor{blue!8}
    \multicolumn{10}{c}{\textit{3B Models}} \\
    \cmidrule(){1-10}
    Qwen2.5-VL-3B  & 77.25  & 48.97 & 46.03 & 65.23 & 59.37 & 82.6 & 84.08 & 68.69  & 78.46 \\
    \rowcolor{gray!10}
    \methodname-3B  & 78.29  & 50.92 & 49.66 & 66.67 &  \textbf{61.39 \textcolor[RGB]{0, 150,0}{(+2.02)}} & 84.1 & 83.08 & 68.56  &  \textbf{78.58 \textcolor[RGB]{0, 150, 0}{(+0.12)}}  \\
    \cmidrule(){1-10} 
    \rowcolor{blue!8}
    \multicolumn{10}{c}{\textit{7B Models}} \\
    \cmidrule(){1-10}
    Qwen2.5-VL-7B  & 82.23 & 55.87 & 52.26 & 68.63 & 64.75 & 88.5 & 86.44& 70.93 & 81.96 \\
    \rowcolor{gray!10}
    \methodname-7B  & 82.60  & 56.23 & 53.18 & 69.28 &  \textbf{65.32 \textcolor[RGB]{0, 150,0}{(+0.57)}} & 89.6 & 88.08 & 71.85  &  \textbf{83.18 \textcolor[RGB]{0, 150, 0}{(+1.22)}}  \\
    \bottomrule
  \end{tabular}
  \vspace{-12pt}
}
  \label{tab:general}
\end{table*}

\subsubsection{General Visual Capabilities}
\label{sec:exp-gen}

\noindent \textbf{Evaluation setup.}
To verify that spatial-focused training does not degrade general visual understanding, we evaluate \methodname on two complementary benchmark suites.
First, we assess general VQA capabilities on $\text{MMBench-v1.1-EN}_\text{test}$~\cite{liu2024mmbench} (general visual understanding), BLINK~\cite{fu2024blink} (multi-image understanding), HallusionBench~\cite{guan2024hallusionbench} (hallucination), and RealWorldQA~\cite{xai_grok_1.5} (real-world scene understanding).
These benchmarks test fundamental visual perception without requiring extensive spatial analysis, so we evaluate both baseline and \methodname models using standard prompts without reasoning instructions.
Second, we evaluate fine-grained visual perception on OCRBench~\cite{Liu_2024}, $\text{ChartQA}_{\text{test}}$~\cite{masry2022chartqabenchmarkquestionanswering}, and SeedBench2-plus~\cite{li2024seedbench2plusbenchmarkingmultimodallarge}, which require accurate recognition of dense text, charts, and detailed visual information.
For these benchmarks, we apply reasoning prompts to \methodname to maintain consistency with spatial evaluation.

\noindent \textbf{General VQA results.} As shown in \cref{tab:spa_v3} and \cref{tab:general}, \methodname not only preserves but actually improves general visual capabilities despite being trained exclusively on self-supervised spatial tasks.
The 3B model achieves an average gain of +2.02\% across the four general VQA benchmarks.
The 7B model shows a smaller but consistent gain of +0.57\% on average, with improvements on MMBench1.1 (+0.37\%), HallusionBench (+0.92\%), and RealWorldQA (+0.65\%).
The consistent gains across diverse visual reasoning tasks suggest that self-supervised spatial training provides a beneficial effect for general visual understanding.
We attribute this to two factors: (1) the pretext tasks require holistic image understanding to reason about patch relationships and structural coherence, which transfers to general scene comprehension; and (2) the reasoning format training encourages systematic visual analysis that benefits non-spatial reasoning as well.

\noindent \textbf{Fine-grained perception results.}
On OCR and chart understanding benchmarks, \methodname maintains strong performance with modest improvements: +0.12\% average for 3B models and +1.22\% average for 7B models.
Notably, the 7B model boosts performance on OCRBench (+1.1\%), ChartQA (+1.64\%), and SeedBench2-plus (+0.92\%).
The minor degradation on ChartQA for the 3B model (-1.0\%) is within typical evaluation variance and does not indicate systematic capability loss.
We attribute the overall preservation to the fact that several pretext tasks, particularly cropped patch inpainting and flipped patch recognition, require attention to fine-grained visual details, texture continuity, and local structural consistency, which align with the demands of OCR and chart understanding.

\subsection{Ablation Studies}\label{sec:ablation}
\begin{table*}[h]
  \caption{\textbf{Task ablation on benchmark subsets.} Each row represents a training configuration and its performance across seven evaluation dimensions. All models are trained based on Qwen2.5-VL-7B. The five columns under \textit{training tasks} illustrate the tasks used for each setting. \textit{Gnr-VQA} averages the four general VQA benchmarks from \cref{tab:general}. Spatial subtasks are tested on the subsets of the spatial benchmarks in \cref{tab:spa}. \textbf{Bold} indicates best performance; \uuline{double-underline} and \uline{underline} indicate second and third best. }
  \vspace{-6pt}
  \centering
  \setlength{\tabcolsep}{2pt}
  \scalebox{0.9}{
  \normalsize
  \begin{tabular}{@{}lcccc|ccccccc@{}}
    \toprule
    \multicolumn{5}{c|}{\textbf{Trained Tasks}} & \multicolumn{7}{c}{\textbf{Benchmark Subsets}} \\
    \cmidrule(){1-5} \cmidrule(){6-12}
    \multicolumn{3}{c}{\textbf{Depth-free}} & \multicolumn{2}{c|}{\textbf{Depth-based}} & \multicolumn{1}{c}{\textbf{General Tasks}} & \multicolumn{2}{c}{\textbf{2D-Spatial Tasks}} & \multicolumn{4}{c}{\textbf{3D-Spatial Tasks}}  \\
\cmidrule(r){1-3} \cmidrule(){4-5} \cmidrule(r){6-6} \cmidrule(r){7-8} \cmidrule(){9-12} 
  {Crop}& {Shuf.} & {Flip} & {Depth} & {Pos.} & Gnr-VQA & Spa457-2D & SpaEval-Maze & Spa457-Pose & 3DSR-Height & 3DSR-Locat. & 3DSR-MultiObj. \\

    \midrule
    \cmidrule(){1-12} 
    \rowcolor{blue!8}
    \multicolumn{12}{c}{\textit{Baseline Model}} \\
    \cmidrule(){1-12}
     \ding{55}& \ding{55} &\ding{55} &\ding{55}&\ding{55} & 64.75  & 56.52 & 35.13 & 33.92 & 52.61 & 67.84  & 46.70  \\
    \cmidrule(){1-12} 
    \rowcolor{blue!8}
    \multicolumn{12}{c}{\textit{Models on Depth-free Tasks}} \\
    \cmidrule(){1-12} 
    \CheckmarkBold &\ding{55} &\ding{55} &\ding{55} &\ding{55} & \uuline{65.39}  & 59.64 & 37.67 & 40.40 & \uline{58.99} & 69.22   & 47.57  \\
    \ding{55} & \CheckmarkBold & \ding{55} &\ding{55} &\ding{55} & 65.30  & 61.30 & \uuline{42.80} & 40.61  & 58.48 & 69.51  & 48.10  \\
     \ding{55}& \ding{55}&\CheckmarkBold & \ding{55} &\ding{55} & \uline{65.38} & 60.86 & \textbf{43.60} & 39.69 & 56.59 & 71.24   & 48.63  \\
     \CheckmarkBold &\CheckmarkBold  &\CheckmarkBold  & \ding{55}& \ding{55}&  \uline{65.38} & \uline{61.68} & 40.80 & \uline{41.36} & 56.67  & 72.06 & 47.79  \\
     \cmidrule(){1-12} 
     \rowcolor{blue!8}
     \multicolumn{12}{c}{\textit{Models on Depth-based Tasks}} \\
    \cmidrule(){1-12} 
      \ding{55}& \ding{55} & \ding{55} &\CheckmarkBold  &\ding{55} & 64.99  & 61.60 & 28.73 & 40.22 & \textbf{63.48}  & \uuline{72.94} & 46.99  \\
     \ding{55}&\ding{55} &\ding{55} &\ding{55} &  \CheckmarkBold& \textbf{65.64}  & 60.86 & 39.07 & 40.20 & 58.91  & \uline{72.20} & \uuline{48.85}  \\
     \ding{55}&\ding{55} &\ding{55} & \CheckmarkBold &\CheckmarkBold  & 65.29& \uuline{62.06} & 36.07 & \uuline{41.67} & \uuline{62.97}  & 71.94 & \textbf{49.28}  \\
          \cmidrule(){1-12}
     \rowcolor{blue!8}
     \multicolumn{12}{c}{\textit{Models on All Tasks}} \\
    \cmidrule(){1-12} 
     \CheckmarkBold&\CheckmarkBold & \CheckmarkBold& \CheckmarkBold &\CheckmarkBold  & 65.32  & \textbf{62.94} & \uline{42.53} & \textbf{42.74} & 58.91  & \textbf{74.11} & \uuline{48.85}  \\
    \bottomrule
  \end{tabular}
  }
  \vspace{-6pt}
  \label{tab:ablation}
\end{table*}

\noindent \textbf{Overall observations.} As shown in \cref{tab:ablation}, all training configurations improve upon the baseline across most evaluation dimensions, validating that each pretext task contributes positively to spatial understanding.
The model trained on \textit{all five tasks} achieves the best performance on four out of seven benchmarks and competitive performance on others, demonstrating that depth-free and depth-based supervision provide complementary learning signals that synergize when combined.
Comparing single-task models to the combined model reveals that no individual task dominates across all dimensions, underscoring the importance of diverse self-supervised objectives for spatial understanding.

\noindent \textbf{Individual task contributions.}
Among depth-free tasks, \textit{Shuffled Patch Reordering} excels at 2D spatial layout and reasoning, likely because recovering patch order requires global structural understanding and multi-step inference.
\textit{Flipped Patch Recognition} achieves the strongest maze reasoning performance, suggesting that detecting subtle orientation violations is beneficial to logical reasoning.
\textit{Cropped Patch Inpainting} shows balanced improvements across general VQA and 3D height estimation, indicating that texture-context matching transfers broadly.
Among depth-based tasks, \textit{Regional Depth Ordering} achieves the strongest 3D height understanding and strong location reasoning, validating its direct supervision of ordinal depth perception.
\textit{Relative Position Prediction} achieves the best general VQA and strong multi-object reasoning, likely because egocentric coordinate transformation requires mental rotation and perspective-taking that transfer to general reasoning.
Combining both depth-based tasks achieves the best multi-object performance, demonstrating complementarity between ordinal depth and egocentric reasoning.

\noindent \textbf{Depth-free vs. depth-based supervision.}
Depth-free tasks drive stronger improvements on general VQA and 2D spatial layout, validating that RGB-only supervision effectively enhances 2D understanding.
Notably, depth-free tasks also improve 3D reasoning (e.g., Flip achieves 48.63\% on 3DSR-MultiObj., competitive with depth-based tasks), suggesting that 2D structural reasoning provides useful inductive biases for 3D inference.
Depth-based tasks, as expected, excel at 3D-centric benchmarks: averaging 61.45\% on the three 3DSR subsets vs. 57.99\% for depth-free tasks.
The 3.46\% gap demonstrates that explicit depth supervision is crucial for robust 3D understanding, though 2D tasks provide non-trivial 3D benefits.

\noindent \textbf{Task complementarity and synergy.}
Combining all five tasks yields improvements beyond any single-category model on key metrics.
This synergy indicates that 2D and 3D supervision address different failure modes and provide mutual regularization.
According to our observations, diverse task coverage is more important than individual task optimization, as no single task suffices.
Practitioners aiming to maximize specific capabilities (e.g., 3D height for robotics) can prioritize corresponding tasks (Regional Depth), while applications requiring holistic spatial intelligence should use diverse task combinations.
\section{Conclusion}
We introduce \methodname, a self-supervised reinforcement learning paradigm that derives verifiable supervision directly from intrinsic image structure.
Our key insight is that ordinary RGB and RGB-D images contain rich self-supervised signals for spatial understanding that are naturally compatible with reinforcement learning through verifiable rewards.
Through comprehensive experiments on seven spatial understanding benchmarks, we demonstrate that \methodname delivers substantial improvements: +4.63\% average accuracy for 3B models and +3.89\% for 7B models, with particularly strong gains on complex spatial reasoning benchmarks.
Critically, \methodname not only enhances spatial capabilities but also improves general visual understanding while preserving fine-grained perception.
In future work, we plan to extend our framework to video-native pretext tasks (e.g., optical flow prediction and temporal coherence) to strengthen video spatial reasoning beyond the current cross-modal transfer.

{
    \small
    \bibliographystyle{ieeenat_fullname}
    \bibliography{main}

\begin{thebibliography}{76}
\providecommand{\natexlab}[1]{#1}
\providecommand{\url}[1]{\texttt{#1}}
\expandafter\ifx\csname urlstyle\endcsname\relax
  \providecommand{\doi}[1]{doi: #1}\else
  \providecommand{\doi}{doi: \begingroup \urlstyle{rm}\Url}\fi

\bibitem[Bai et~al.(2025)Bai, Chen, Liu, Wang, Ge, Song, Dang, Wang, Wang, Tang, et~al.]{bai2025qwen2}
Shuai Bai, Keqin Chen, Xuejing Liu, Jialin Wang, Wenbin Ge, Sibo Song, Kai Dang, Peng Wang, Shijie Wang, Jun Tang, et~al.
\newblock Qwen2. 5-vl technical report.
\newblock \emph{arXiv preprint arXiv:2502.13923}, 2025.

\bibitem[Brown et~al.(2020)Brown, Mann, Ryder, Subbiah, Kaplan, Dhariwal, Neelakantan, Shyam, Sastry, Askell, et~al.]{brown2020languagemodelsfewshotlearners}
Tom Brown, Benjamin Mann, Nick Ryder, Melanie Subbiah, Jared~D Kaplan, Prafulla Dhariwal, Arvind Neelakantan, Pranav Shyam, Girish Sastry, Amanda Askell, et~al.
\newblock Language models are few-shot learners.
\newblock \emph{Advances in neural information processing systems}, 33:\penalty0 1877--1901, 2020.

\bibitem[Cai et~al.(2025)Cai, Ponomarenko, Yuan, Li, Yang, Dong, and Zhao]{cai2025spatialbotprecisespatialunderstanding}
Wenxiao Cai, Iaroslav Ponomarenko, Jianhao Yuan, Xiaoqi Li, Wankou Yang, Hao Dong, and Bo Zhao.
\newblock Spatialbot: Precise spatial understanding with vision language models.
\newblock In \emph{2025 IEEE International Conference on Robotics and Automation (ICRA)}, pages 9490--9498. IEEE, 2025.

\bibitem[Chen et~al.(2024{\natexlab{a}})Chen, Xu, Kirmani, Ichter, Sadigh, Guibas, and Xia]{chen2024spatialvlm}
Boyuan Chen, Zhuo Xu, Sean Kirmani, Brain Ichter, Dorsa Sadigh, Leonidas Guibas, and Fei Xia.
\newblock {SpatialVLM}: Endowing vision-language models with spatial reasoning capabilities.
\newblock In \emph{CVPR}, 2024{\natexlab{a}}.

\bibitem[Chen et~al.(2025)Chen, Tu, Wang, Liu, Tang, Du, Zhou, and Xie]{chen2025sftrlearlyinvestigation}
Hardy Chen, Haoqin Tu, Fali Wang, Hui Liu, Xianfeng Tang, Xinya Du, Yuyin Zhou, and Cihang Xie.
\newblock Sft or rl? an early investigation into training r1-like reasoning large vision-language models.
\newblock \emph{arXiv preprint arXiv:2504.11468}, 2025.

\bibitem[Chen et~al.(2024{\natexlab{b}})Chen, Li, Dong, Zhang, Zang, Chen, Duan, Wang, Qiao, Lin, et~al.]{chen2024we}
Lin Chen, Jinsong Li, Xiaoyi Dong, Pan Zhang, Yuhang Zang, Zehui Chen, Haodong Duan, Jiaqi Wang, Yu Qiao, Dahua Lin, et~al.
\newblock Are we on the right way for evaluating large vision-language models?
\newblock In \emph{NeurIPs}, 2024{\natexlab{b}}.

\bibitem[Chen et~al.(2020)Chen, Kornblith, Norouzi, and Hinton]{chen2020simpleframeworkcontrastivelearning}
Ting Chen, Simon Kornblith, Mohammad Norouzi, and Geoffrey Hinton.
\newblock A simple framework for contrastive learning of visual representations.
\newblock In \emph{International conference on machine learning}, pages 1597--1607. PmLR, 2020.

\bibitem[Cheng et~al.(2024{\natexlab{a}})Cheng, Ji, Yang, Gongye, Zou, Kautz, B{\i}y{\i}k, Yin, Liu, and Wang]{cheng2025navilaleggedrobotvisionlanguageaction}
An-Chieh Cheng, Yandong Ji, Zhaojing Yang, Zaitian Gongye, Xueyan Zou, Jan Kautz, Erdem B{\i}y{\i}k, Hongxu Yin, Sifei Liu, and Xiaolong Wang.
\newblock Navila: Legged robot vision-language-action model for navigation.
\newblock \emph{arXiv preprint arXiv:2412.04453}, 2024{\natexlab{a}}.

\bibitem[Cheng et~al.(2024{\natexlab{b}})Cheng, Yin, Fu, Guo, Yang, Kautz, Wang, and Liu]{cheng2024spatialrgpt}
An-Chieh Cheng, Hongxu Yin, Yang Fu, Qiushan Guo, Ruihan Yang, Jan Kautz, Xiaolong Wang, and Sifei Liu.
\newblock {SpatialRGPT}: Grounded spatial reasoning in vision-language models.
\newblock In \emph{NeurIPs}, 2024{\natexlab{b}}.

\bibitem[Chu et~al.(2025)Chu, Zhai, Yang, Tong, Xie, Schuurmans, Le, Levine, and Ma]{chu2025sftmemorizesrlgeneralizes}
Tianzhe Chu, Yuexiang Zhai, Jihan Yang, Shengbang Tong, Saining Xie, Dale Schuurmans, Quoc~V Le, Sergey Levine, and Yi Ma.
\newblock Sft memorizes, rl generalizes: A comparative study of foundation model post-training.
\newblock \emph{arXiv preprint arXiv:2501.17161}, 2025.

\bibitem[Dai et~al.(2017)Dai, Chang, Savva, Halber, Funkhouser, and Nie{\ss}ner]{dai2017scannet}
Angela Dai, Angel~X Chang, Manolis Savva, Maciej Halber, Thomas Funkhouser, and Matthias Nie{\ss}ner.
\newblock Scannet: Richly-annotated 3d reconstructions of indoor scenes.
\newblock In \emph{Proceedings of the IEEE conference on computer vision and pattern recognition}, pages 5828--5839, 2017.

\bibitem[Deng et~al.(2025)Deng, Gu, Ye, He, Chen, Li, Wang, Wei, Yang, Dou, et~al.]{deng2025internspatialcomprehensivedatasetspatial}
Nianchen Deng, Lixin Gu, Shenglong Ye, Yinan He, Zhe Chen, Songze Li, Haomin Wang, Xingguang Wei, Tianshuo Yang, Min Dou, et~al.
\newblock Internspatial: A comprehensive dataset for spatial reasoning in vision-language models.
\newblock \emph{arXiv preprint arXiv:2506.18385}, 2025.

\bibitem[Devlin et~al.(2019)Devlin, Chang, Lee, and Toutanova]{devlin2019bertpretrainingdeepbidirectional}
Jacob Devlin, Ming-Wei Chang, Kenton Lee, and Kristina Toutanova.
\newblock Bert: Pre-training of deep bidirectional transformers for language understanding.
\newblock In \emph{Proceedings of the 2019 conference of the North American chapter of the association for computational linguistics: human language technologies, volume 1 (long and short papers)}, pages 4171--4186, 2019.

\bibitem[Driess et~al.(2023)Driess, Xia, Sajjadi, Lynch, Chowdhery, Wahid, Tompson, Vuong, Yu, Huang, et~al.]{driess2023palmeembodiedmultimodallanguage}
Danny Driess, Fei Xia, Mehdi~SM Sajjadi, Corey Lynch, Aakanksha Chowdhery, Ayzaan Wahid, Jonathan Tompson, Quan Vuong, Tianhe Yu, Wenlong Huang, et~al.
\newblock Palm-e: An embodied multimodal language model.
\newblock In \emph{arXiv preprint arXiv:2303.03378}, 2023.

\bibitem[Duan et~al.(2024)Duan, Yang, Qiao, Fang, Chen, Liu, Dong, Zang, Zhang, Wang, et~al.]{duan2025vlmevalkitopensourcetoolkitevaluating}
Haodong Duan, Junming Yang, Yuxuan Qiao, Xinyu Fang, Lin Chen, Yuan Liu, Xiaoyi Dong, Yuhang Zang, Pan Zhang, Jiaqi Wang, et~al.
\newblock Vlmevalkit: An open-source toolkit for evaluating large multi-modality models.
\newblock In \emph{Proceedings of the 32nd ACM international conference on multimedia}, pages 11198--11201, 2024.

\bibitem[Feng et~al.(2025)Feng, Liu, Yang, Cai, Zhang, Zhan, Huang, Yan, Wan, Liu, et~al.]{feng2025vision}
Yongchao Feng, Yajie Liu, Shuai Yang, Wenrui Cai, Jinqing Zhang, Qiqi Zhan, Ziyue Huang, Hongxi Yan, Qiao Wan, Chenguang Liu, et~al.
\newblock Vision-language model for object detection and segmentation: A review and evaluation.
\newblock \emph{arXiv preprint arXiv:2504.09480}, 2025.

\bibitem[Fu et~al.(2024)Fu, Hu, Li, Feng, Wang, Lin, Roth, Smith, Ma, and Krishna]{fu2024blink}
Xingyu Fu, Yushi Hu, Bangzheng Li, Yu Feng, Haoyu Wang, Xudong Lin, Dan Roth, Noah~A Smith, Wei-Chiu Ma, and Ranjay Krishna.
\newblock Blink: Multimodal large language models can see but not perceive.
\newblock In \emph{European Conference on Computer Vision}, pages 148--166. Springer, 2024.

\bibitem[Gao et~al.(2024)Gao, Xu, Ye, Liu, He, Fu, Mei, Wang, and Wu]{gao2024designing}
Jiaxuan Gao, Shusheng Xu, Wenjie Ye, Weilin Liu, Chuyi He, Wei Fu, Zhiyu Mei, Guangju Wang, and Yi Wu.
\newblock On designing effective rl reward at training time for llm reasoning.
\newblock \emph{arXiv preprint arXiv:2410.15115}, 2024.

\bibitem[Gidaris et~al.(2018)Gidaris, Singh, and Komodakis]{gidaris2018unsupervisedrepresentationlearningpredicting}
Spyros Gidaris, Praveer Singh, and Nikos Komodakis.
\newblock Unsupervised representation learning by predicting image rotations.
\newblock \emph{arXiv preprint arXiv:1803.07728}, 2018.

\bibitem[Goetting et~al.(2024)Goetting, Singh, and Loquercio]{goetting2024endtoendnavigationvisionlanguage}
Dylan Goetting, Himanshu~Gaurav Singh, and Antonio Loquercio.
\newblock End-to-end navigation with vision language models: Transforming spatial reasoning into question-answering.
\newblock \emph{arXiv preprint arXiv:2411.05755}, 2024.

\bibitem[Guan et~al.(2024)Guan, Liu, Wu, Xian, Li, Liu, Wang, Chen, Huang, Yacoob, et~al.]{guan2024hallusionbench}
Tianrui Guan, Fuxiao Liu, Xiyang Wu, Ruiqi Xian, Zongxia Li, Xiaoyu Liu, Xijun Wang, Lichang Chen, Furong Huang, Yaser Yacoob, et~al.
\newblock Hallusionbench: an advanced diagnostic suite for entangled language hallucination and visual illusion in large vision-language models.
\newblock In \emph{Proceedings of the IEEE/CVF Conference on Computer Vision and Pattern Recognition}, pages 14375--14385, 2024.

\bibitem[Gui et~al.(2024)Gui, Chen, Zhang, Cao, Sun, Luo, and Tao]{gui2024surveyselfsupervisedlearningalgorithms}
Jie Gui, Tuo Chen, Jing Zhang, Qiong Cao, Zhenan Sun, Hao Luo, and Dacheng Tao.
\newblock A survey on self-supervised learning: Algorithms, applications, and future trends.
\newblock \emph{IEEE Transactions on Pattern Analysis and Machine Intelligence}, 46\penalty0 (12):\penalty0 9052--9071, 2024.

\bibitem[Guo et~al.(2025{\natexlab{a}})Guo, Yang, Zhang, Song, Zhang, Xu, Zhu, Ma, Wang, Bi, et~al.]{guo2025deepseek}
Daya Guo, Dejian Yang, Haowei Zhang, Junxiao Song, Ruoyu Zhang, Runxin Xu, Qihao Zhu, Shirong Ma, Peiyi Wang, Xiao Bi, et~al.
\newblock Deepseek-r1: Incentivizing reasoning capability in llms via reinforcement learning.
\newblock \emph{arXiv preprint arXiv:2501.12948}, 2025{\natexlab{a}}.

\bibitem[Guo et~al.(2025{\natexlab{b}})Guo, Zhou, Wang, Zhang, Zhang, Jegelka, Wang, Chai, Yin, Lin, et~al.]{guo2025ssl4rlrevisitingselfsupervisedlearning}
Xiaojun Guo, Runyu Zhou, Yifei Wang, Qi Zhang, Chenheng Zhang, Stefanie Jegelka, Xiaohan Wang, Jiajun Chai, Guojun Yin, Wei Lin, et~al.
\newblock Ssl4rl: Revisiting self-supervised learning as intrinsic reward for visual-language reasoning.
\newblock \emph{arXiv preprint arXiv:2510.16416}, 2025{\natexlab{b}}.

\bibitem[He et~al.(2020)He, Fan, Wu, Xie, and Girshick]{he2020momentum}
Kaiming He, Haoqi Fan, Yuxin Wu, Saining Xie, and Ross Girshick.
\newblock Momentum contrast for unsupervised visual representation learning.
\newblock In \emph{CVPR}, 2020.

\bibitem[He et~al.(2022)He, Chen, Xie, Li, Doll{\'a}r, and Girshick]{he2021maskedautoencodersscalablevision}
Kaiming He, Xinlei Chen, Saining Xie, Yanghao Li, Piotr Doll{\'a}r, and Ross Girshick.
\newblock Masked autoencoders are scalable vision learners.
\newblock In \emph{Proceedings of the IEEE/CVF conference on computer vision and pattern recognition}, pages 16000--16009, 2022.

\bibitem[Hong et~al.(2023)Hong, Lin, Du, Chen, Tenenbaum, and Gan]{hong20233dconceptlearningreasoning}
Yining Hong, Chunru Lin, Yilun Du, Zhenfang Chen, Joshua~B Tenenbaum, and Chuang Gan.
\newblock 3d concept learning and reasoning from multi-view images.
\newblock In \emph{Proceedings of the IEEE/CVF Conference on Computer Vision and Pattern Recognition}, pages 9202--9212, 2023.

\bibitem[Kamath et~al.(2023)Kamath, Hessel, and Chang]{kamath2023whatsupvisionlanguagemodels}
Amita Kamath, Jack Hessel, and Kai-Wei Chang.
\newblock What's" up" with vision-language models? investigating their struggle with spatial reasoning.
\newblock \emph{arXiv preprint arXiv:2310.19785}, 2023.

\bibitem[Kim et~al.(2024)Kim, Pertsch, Karamcheti, Xiao, Balakrishna, Nair, Rafailov, Foster, Lam, Sanketi, et~al.]{kim2024openvlaopensourcevisionlanguageactionmodel}
Moo~Jin Kim, Karl Pertsch, Siddharth Karamcheti, Ted Xiao, Ashwin Balakrishna, Suraj Nair, Rafael Rafailov, Ethan Foster, Grace Lam, Pannag Sanketi, et~al.
\newblock Openvla: An open-source vision-language-action model.
\newblock \emph{arXiv preprint arXiv:2406.09246}, 2024.

\bibitem[Lambert et~al.(2024)Lambert, Morrison, Pyatkin, Huang, Ivison, Brahman, Miranda, Liu, Dziri, Lyu, et~al.]{lambert2025tulu3pushingfrontiers}
Nathan Lambert, Jacob Morrison, Valentina Pyatkin, Shengyi Huang, Hamish Ivison, Faeze Brahman, Lester James~V Miranda, Alisa Liu, Nouha Dziri, Shane Lyu, et~al.
\newblock Tulu 3: Pushing frontiers in open language model post-training.
\newblock \emph{arXiv preprint arXiv:2411.15124}, 2024.

\bibitem[Li et~al.(2024)Li, Ge, Chen, Ge, Zhang, and Shan]{li2024seedbench2plusbenchmarkingmultimodallarge}
Bohao Li, Yuying Ge, Yi Chen, Yixiao Ge, Ruimao Zhang, and Ying Shan.
\newblock Seed-bench-2-plus: Benchmarking multimodal large language models with text-rich visual comprehension.
\newblock \emph{arXiv preprint arXiv:2404.16790}, 2024.

\bibitem[Li et~al.(2025{\natexlab{a}})Li, Li, Wang, Yan, Zhang, Chen, Hou, Jiang, Zhang, Shen, et~al.]{li2025viewspatialbenchevaluatingmultiperspectivespatial}
Dingming Li, Hongxing Li, Zixuan Wang, Yuchen Yan, Hang Zhang, Siqi Chen, Guiyang Hou, Shengpei Jiang, Wenqi Zhang, Yongliang Shen, et~al.
\newblock Viewspatial-bench: Evaluating multi-perspective spatial localization in vision-language models.
\newblock \emph{arXiv preprint arXiv:2505.21500}, 2025{\natexlab{a}}.

\bibitem[Li et~al.(2025{\natexlab{b}})Li, Li, Wang, Yan, Wu, Zhang, Shen, Lu, Xiao, and Zhuang]{li2025spatialladderprogressivetrainingspatial}
Hongxing Li, Dingming Li, Zixuan Wang, Yuchen Yan, Hang Wu, Wenqi Zhang, Yongliang Shen, Weiming Lu, Jun Xiao, and Yueting Zhuang.
\newblock Spatialladder: Progressive training for spatial reasoning in vision-language models.
\newblock \emph{arXiv preprint arXiv:2510.08531}, 2025{\natexlab{b}}.

\bibitem[Li et~al.(2025{\natexlab{c}})Li, Zhang, Lin, Liu, Cai, Liu, and Zhao]{li2025stibenchmllmsreadyprecise}
Yun Li, Yiming Zhang, Tao Lin, XiangRui Liu, Wenxiao Cai, Zheng Liu, and Bo Zhao.
\newblock Sti-bench: Are mllms ready for precise spatial-temporal world understanding?
\newblock \emph{arXiv preprint arXiv:2503.23765}, 2025{\natexlab{c}}.

\bibitem[Li and Snavely(2018)]{li2018megadepth}
Zhengqi Li and Noah Snavely.
\newblock Megadepth: Learning single-view depth prediction from internet photos.
\newblock In \emph{Proceedings of the IEEE conference on computer vision and pattern recognition}, pages 2041--2050, 2018.

\bibitem[Liao et~al.(2024)Liao, Mahmood, Fidler, and Acuna]{liao2024reasoningpathsreferenceobjects}
Yuan-Hong Liao, Rafid Mahmood, Sanja Fidler, and David Acuna.
\newblock Reasoning paths with reference objects elicit quantitative spatial reasoning in large vision-language models.
\newblock \emph{arXiv preprint arXiv:2409.09788}, 2024.

\bibitem[Liao et~al.(2025)Liao, Xie, Zhang, Kong, Lu, Yang, and Deng]{liao2025improved}
Zhenyi Liao, Qingsong Xie, Yanhao Zhang, Zijian Kong, Haonan Lu, Zhenyu Yang, and Zhijie Deng.
\newblock Improved visual-spatial reasoning via r1-zero-like training.
\newblock \emph{arXiv preprint arXiv:2504.00883}, 2025.

\bibitem[Lin et~al.(2014)Lin, Maire, Belongie, Hays, Perona, Ramanan, Doll{\'a}r, and Zitnick]{lin2014microsoft}
Tsung-Yi Lin, Michael Maire, Serge Belongie, James Hays, Pietro Perona, Deva Ramanan, Piotr Doll{\'a}r, and C~Lawrence Zitnick.
\newblock Microsoft coco: Common objects in context.
\newblock In \emph{European conference on computer vision}, pages 740--755. Springer, 2014.

\bibitem[Liu et~al.(2024{\natexlab{a}})Liu, Duan, Zhang, Li, Zhang, Zhao, Yuan, Wang, He, Liu, et~al.]{liu2024mmbench}
Yuan Liu, Haodong Duan, Yuanhan Zhang, Bo Li, Songyang Zhang, Wangbo Zhao, Yike Yuan, Jiaqi Wang, Conghui He, Ziwei Liu, et~al.
\newblock {MMBench}: Is your multi-modal model an all-around player?
\newblock In \emph{ECCV}, 2024{\natexlab{a}}.

\bibitem[Liu et~al.(2024{\natexlab{b}})Liu, Li, Huang, Yang, Yu, Li, Yin, Liu, Jin, and Bai]{Liu_2024}
Yuliang Liu, Zhang Li, Mingxin Huang, Biao Yang, Wenwen Yu, Chunyuan Li, Xu-Cheng Yin, Cheng-Lin Liu, Lianwen Jin, and Xiang Bai.
\newblock Ocrbench: on the hidden mystery of ocr in large multimodal models.
\newblock \emph{Science China Information Sciences}, 67\penalty0 (12), 2024{\natexlab{b}}.

\bibitem[Liu et~al.(2025{\natexlab{a}})Liu, Niu, Xiao, Zheng, Yuan, Zang, Cao, Dong, Liang, Chen, Sun, Lin, and Wang]{liu2025starbench}
Zihan Liu, Zhikang Niu, Qiuyang Xiao, Zhisheng Zheng, Ruoqi Yuan, Yuhang Zang, Yuhang Cao, Xiaoyi Dong, Jianze Liang, Xie Chen, Leilei Sun, Dahua Lin, and Jiaqi Wang.
\newblock Star-bench: Probing deep spatio-temporal reasoning as audio 4d intelligence.
\newblock \emph{arXiv preprint arXiv:2510.24693}, 2025{\natexlab{a}}.

\bibitem[Liu et~al.(2025{\natexlab{b}})Liu, Sun, Zang, Dong, Cao, Duan, Lin, and Wang]{liu2025visual}
Ziyu Liu, Zeyi Sun, Yuhang Zang, Xiaoyi Dong, Yuhang Cao, Haodong Duan, Dahua Lin, and Jiaqi Wang.
\newblock {Visual-RFT}: Visual reinforcement fine-tuning.
\newblock In \emph{ICCV}, 2025{\natexlab{b}}.

\bibitem[Lu et~al.(2023)Lu, Bansal, Xia, Liu, Li, Hajishirzi, Cheng, Chang, Galley, and Gao]{lu2023mathvista}
Pan Lu, Hritik Bansal, Tony Xia, Jiacheng Liu, Chunyuan Li, Hannaneh Hajishirzi, Hao Cheng, Kai-Wei Chang, Michel Galley, and Jianfeng Gao.
\newblock {MathVista}: Evaluating mathematical reasoning of foundation models in visual contexts.
\newblock \emph{arXiv preprint arXiv:2310.02255}, 2023.

\bibitem[Ma et~al.(2025)Ma, Chen, Zhang, Chou, Chen, de~Melo, and Yuille]{ma20253dsrbenchcomprehensive3dspatial}
Wufei Ma, Haoyu Chen, Guofeng Zhang, Yu-Cheng Chou, Jieneng Chen, Celso de Melo, and Alan Yuille.
\newblock 3dsrbench: A comprehensive 3d spatial reasoning benchmark.
\newblock In \emph{Proceedings of the IEEE/CVF International Conference on Computer Vision}, pages 6924--6934, 2025.

\bibitem[Masry et~al.(2022)Masry, Long, Tan, Joty, and Hoque]{masry2022chartqabenchmarkquestionanswering}
Ahmed Masry, Do~Xuan Long, Jia~Qing Tan, Shafiq Joty, and Enamul Hoque.
\newblock Chartqa: A benchmark for question answering about charts with visual and logical reasoning.
\newblock \emph{arXiv preprint arXiv:2203.10244}, 2022.

\bibitem[Misra et~al.(2016)Misra, Zitnick, and Hebert]{misra2016shufflelearnunsupervisedlearning}
Ishan Misra, C~Lawrence Zitnick, and Martial Hebert.
\newblock Shuffle and learn: unsupervised learning using temporal order verification.
\newblock In \emph{European conference on computer vision}, pages 527--544. Springer, 2016.

\bibitem[Noroozi and Favaro(2016)]{noroozi2017unsupervisedlearningvisualrepresentations}
Mehdi Noroozi and Paolo Favaro.
\newblock Unsupervised learning of visual representations by solving jigsaw puzzles.
\newblock In \emph{European conference on computer vision}, pages 69--84. Springer, 2016.

\bibitem[Ouyang et~al.(2025)Ouyang, Liu, Wu, Liu, Zhou, Zhou, Meng, and Sun]{ouyang2025spacerreinforcingmllmsvideo}
Kun Ouyang, Yuanxin Liu, Haoning Wu, Yi Liu, Hao Zhou, Jie Zhou, Fandong Meng, and Xu Sun.
\newblock Spacer: Reinforcing mllms in video spatial reasoning.
\newblock \emph{arXiv preprint arXiv:2504.01805}, 2025.

\bibitem[Qi et~al.(2025)Qi, Zhang, Fang, Wang, and Zhao]{qi2025gpt4scene}
Zhangyang Qi, Zhixiong Zhang, Ye Fang, Jiaqi Wang, and Hengshuang Zhao.
\newblock Gpt4scene: Understand 3d scenes from videos with vision-language models.
\newblock \emph{arXiv preprint arXiv:2501.01428}, 2025.

\bibitem[Ro et~al.(2025)Ro, Kim, and Yoon]{ro2025visionlanguagemodelsunderstandcities}
Juneyoung Ro, Namwoo Kim, and Yoonjin Yoon.
\newblock How well do vision-language models understand cities? a comparative study on spatial reasoning from street-view images.
\newblock In \emph{Proceedings of the IEEE/CVF International Conference on Computer Vision}, pages 6476--6485, 2025.

\bibitem[Santa~Cruz et~al.(2017)Santa~Cruz, Fernando, Cherian, and Gould]{cruz2017deeppermnetvisualpermutationlearning}
Rodrigo Santa~Cruz, Basura Fernando, Anoop Cherian, and Stephen Gould.
\newblock Deeppermnet: Visual permutation learning.
\newblock In \emph{Proceedings of the IEEE Conference on Computer Vision and Pattern Recognition}, pages 3949--3957, 2017.

\bibitem[Sarbolandi et~al.(2015)Sarbolandi, Lefloch, and Kolb]{sarbolandi2015kinect}
Hamed Sarbolandi, Damien Lefloch, and Andreas Kolb.
\newblock Kinect range sensing: Structured-light versus time-of-flight kinect.
\newblock \emph{Computer vision and image understanding}, 139:\penalty0 1--20, 2015.

\bibitem[Shao et~al.(2024)Shao, Wang, Zhu, Xu, Song, Bi, Zhang, Zhang, Li, Wu, et~al.]{shao2024deepseekmath}
Zhihong Shao, Peiyi Wang, Qihao Zhu, Runxin Xu, Junxiao Song, Xiao Bi, Haowei Zhang, Mingchuan Zhang, YK Li, Yang Wu, et~al.
\newblock Deepseekmath: Pushing the limits of mathematical reasoning in open language models.
\newblock \emph{arXiv preprint arXiv:2402.03300}, 2024.

\bibitem[Song et~al.(2025)Song, Blukis, Tremblay, Tyree, Su, and Birchfield]{song2025robospatialteachingspatialunderstanding}
Chan~Hee Song, Valts Blukis, Jonathan Tremblay, Stephen Tyree, Yu Su, and Stan Birchfield.
\newblock Robospatial: Teaching spatial understanding to 2d and 3d vision-language models for robotics.
\newblock In \emph{Proceedings of the Computer Vision and Pattern Recognition Conference}, pages 15768--15780, 2025.

\bibitem[Tian et~al.(2025)Tian, Mao, Zhang, Jiang, Zhou, and Tu]{tian2025nuscenes}
Kexin Tian, Jingrui Mao, Yunlong Zhang, Jiwan Jiang, Yang Zhou, and Zhengzhong Tu.
\newblock Nuscenes-spatialqa: A spatial understanding and reasoning benchmark for vision-language models in autonomous driving.
\newblock \emph{arXiv preprint arXiv:2504.03164}, 2025.

\bibitem[Vasiljevic et~al.(2019)Vasiljevic, Kolkin, Zhang, Luo, Wang, Dai, Daniele, Mostajabi, Basart, Walter, et~al.]{vasiljevic2019diode}
Igor Vasiljevic, Nick Kolkin, Shanyi Zhang, Ruotian Luo, Haochen Wang, Falcon~Z Dai, Andrea~F Daniele, Mohammadreza Mostajabi, Steven Basart, Matthew~R Walter, et~al.
\newblock Diode: A dense indoor and outdoor depth dataset.
\newblock \emph{arXiv preprint arXiv:1908.00463}, 2019.

\bibitem[Vaswani et~al.(2017)Vaswani, Shazeer, Parmar, Uszkoreit, Jones, Gomez, Kaiser, and Polosukhin]{vaswani2017attention}
Ashish Vaswani, Noam Shazeer, Niki Parmar, Jakob Uszkoreit, Llion Jones, Aidan~N Gomez, {\L}ukasz Kaiser, and Illia Polosukhin.
\newblock Attention is all you need.
\newblock In \emph{NeurIPS}, 2017.

\bibitem[Wang et~al.(2024)Wang, Ming, Shi, Vineet, Wang, Li, and Joshi]{wang2024pictureworththousandwords}
Jiayu Wang, Yifei Ming, Zhenmei Shi, Vibhav Vineet, Xin Wang, Sharon Li, and Neel Joshi.
\newblock Is a picture worth a thousand words? delving into spatial reasoning for vision language models.
\newblock \emph{Advances in Neural Information Processing Systems}, 37:\penalty0 75392--75421, 2024.

\bibitem[Wang et~al.(2025{\natexlab{a}})Wang, Yu, Jiang, Lan, Shi, Chang, Kautz, Li, and Alvarez]{wang2025omnidriveholisticvisionlanguagedataset}
Shihao Wang, Zhiding Yu, Xiaohui Jiang, Shiyi Lan, Min Shi, Nadine Chang, Jan Kautz, Ying Li, and Jose~M Alvarez.
\newblock Omnidrive: A holistic vision-language dataset for autonomous driving with counterfactual reasoning.
\newblock In \emph{Proceedings of the Computer Vision and Pattern Recognition Conference}, pages 22442--22452, 2025{\natexlab{a}}.

\bibitem[Wang et~al.(2025{\natexlab{b}})Wang, Ma, Zhang, de~Melo, Chen, and Yuille]{wang2025spatial457diagnosticbenchmark6d}
Xingrui Wang, Wufei Ma, Tiezheng Zhang, Celso~M de Melo, Jieneng Chen, and Alan Yuille.
\newblock Spatial457: A diagnostic benchmark for 6d spatial reasoning of large mutimodal models.
\newblock In \emph{Proceedings of the Computer Vision and Pattern Recognition Conference}, pages 24669--24679, 2025{\natexlab{b}}.

\bibitem[Wang et~al.(2025{\natexlab{c}})Wang, Yang, Zeng, Ren, Liu, Peng, Cheng, He, Wang, Gao, et~al.]{wang2025reinforcement}
Yiping Wang, Qing Yang, Zhiyuan Zeng, Liliang Ren, Liyuan Liu, Baolin Peng, Hao Cheng, Xuehai He, Kuan Wang, Jianfeng Gao, et~al.
\newblock Reinforcement learning for reasoning in large language models with one training example.
\newblock \emph{arXiv preprint arXiv:2504.20571}, 2025{\natexlab{c}}.

\bibitem[Wang et~al.(2025{\natexlab{d}})Wang, Zhu, Tang, Li, Xiong, Yu, and Blaschko]{wang2025jigsaw}
Zifu Wang, Junyi Zhu, Bo Tang, Zhiyu Li, Feiyu Xiong, Jiaqian Yu, and Matthew~B Blaschko.
\newblock Jigsaw-r1: A study of rule-based visual reinforcement learning with jigsaw puzzles.
\newblock \emph{arXiv preprint arXiv:2505.23590}, 2025{\natexlab{d}}.

\bibitem[Wen et~al.(2025)Wen, Liu, Zheng, Xu, Ye, Wu, Liang, Wang, Li, Miao, et~al.]{wen2025reinforcement}
Xumeng Wen, Zihan Liu, Shun Zheng, Zhijian Xu, Shengyu Ye, Zhirong Wu, Xiao Liang, Yang Wang, Junjie Li, Ziming Miao, et~al.
\newblock Reinforcement learning with verifiable rewards implicitly incentivizes correct reasoning in base llms.
\newblock \emph{arXiv preprint arXiv:2506.14245}, 2025.

\bibitem[Wu et~al.(2025{\natexlab{a}})Wu, Guan, Feng, Liu, Wu, Wang, Wu, and Tan]{wu2025reinforcingspatialreasoningvisionlanguage}
Junfei Wu, Jian Guan, Kaituo Feng, Qiang Liu, Shu Wu, Liang Wang, Wei Wu, and Tieniu Tan.
\newblock Reinforcing spatial reasoning in vision-language models with interwoven thinking and visual drawing.
\newblock \emph{arXiv preprint arXiv:2506.09965}, 2025{\natexlab{a}}.

\bibitem[Wu et~al.(2025{\natexlab{b}})Wu, Zhang, Diao, Li, Lu, and Liu]{wu2025visualjigsawposttrainingimproves}
Penghao Wu, Yushan Zhang, Haiwen Diao, Bo Li, Lewei Lu, and Ziwei Liu.
\newblock Visual jigsaw post-training improves mllms.
\newblock \emph{arXiv preprint arXiv:2509.25190}, 2025{\natexlab{b}}.

\bibitem[X.AI(2024)]{xai_grok_1.5}
X.AI.
\newblock Grok-1.5 vision preview, 2024.

\bibitem[Xing et~al.(2025)Xing, Dong, Zang, Cao, Liang, Huang, Wang, Wu, and Lin]{xing2025caprlstimulatingdenseimage}
Long Xing, Xiaoyi Dong, Yuhang Zang, Yuhang Cao, Jianze Liang, Qidong Huang, Jiaqi Wang, Feng Wu, and Dahua Lin.
\newblock Caprl: Stimulating dense image caption capabilities via reinforcement learning.
\newblock \emph{arXiv preprint arXiv:2509.22647}, 2025.

\bibitem[Yang et~al.(2025)Yang, Yang, Gupta, Han, Fei-Fei, and Xie]{yang2025thinkingspacemultimodallarge}
Jihan Yang, Shusheng Yang, Anjali~W Gupta, Rilyn Han, Li Fei-Fei, and Saining Xie.
\newblock Thinking in space: How multimodal large language models see, remember, and recall spaces.
\newblock In \emph{Proceedings of the Computer Vision and Pattern Recognition Conference}, pages 10632--10643, 2025.

\bibitem[Yu et~al.(2024)Yu, Sun, Zhang, Cui, Zhang, Cao, Wang, and Liu]{yu2024capsfusionrethinkingimagetextdata}
Qiying Yu, Quan Sun, Xiaosong Zhang, Yufeng Cui, Fan Zhang, Yue Cao, Xinlong Wang, and Jingjing Liu.
\newblock Capsfusion: Rethinking image-text data at scale.
\newblock In \emph{Proceedings of the IEEE/CVF Conference on Computer Vision and Pattern Recognition}, pages 14022--14032, 2024.

\bibitem[Zha et~al.(2025)Zha, Fan, Yang, Gao, and Chen]{zha2025enablellm3dcapacity}
Jirong Zha, Yuxuan Fan, Xiao Yang, Chen Gao, and Xinlei Chen.
\newblock How to enable llm with 3d capacity? a survey of spatial reasoning in llm.
\newblock \emph{arXiv preprint arXiv:2504.05786}, 2025.

\bibitem[Zhang et~al.(2025{\natexlab{a}})Zhang, Liu, Dong, Zang, Zhang, Duan, Cao, Lin, and Wang]{zhang2025booststep}
Beichen Zhang, Yuhong Liu, Xiaoyi Dong, Yuhang Zang, Pan Zhang, Haodong Duan, Yuhang Cao, Dahua Lin, and Jiaqi Wang.
\newblock Booststep: Boosting mathematical capability of large language models via improved single-step reasoning.
\newblock \emph{arXiv preprint arXiv:2501.03226}, 2025{\natexlab{a}}.

\bibitem[Zhang et~al.(2025{\natexlab{b}})Zhang, Liu, Li, Wen, Guan, Wang, and Nie]{zhang2025spatialunderstandingvideosstructured}
Haoyu Zhang, Meng Liu, Zaijing Li, Haokun Wen, Weili Guan, Yaowei Wang, and Liqiang Nie.
\newblock Spatial understanding from videos: Structured prompts meet simulation data.
\newblock \emph{arXiv preprint arXiv:2506.03642}, 2025{\natexlab{b}}.

\bibitem[Zhang et~al.(2025{\natexlab{c}})Zhang, Wu, Zhang, Zhao, and Bian]{zhang2025right}
Qingyang Zhang, Haitao Wu, Changqing Zhang, Peilin Zhao, and Yatao Bian.
\newblock Right question is already half the answer: Fully unsupervised llm reasoning incentivization.
\newblock \emph{arXiv preprint arXiv:2504.05812}, 2025{\natexlab{c}}.

\bibitem[Zhang et~al.(2025{\natexlab{d}})Zhang, Huang, Xu, Huang, Zhi, Ren, Xu, and Zhang]{zhang2025mllmsstrugglespatialunderstanding}
Wanyue Zhang, Yibin Huang, Yangbin Xu, JingJing Huang, Helu Zhi, Shuo Ren, Wang Xu, and Jiajun Zhang.
\newblock Why do mllms struggle with spatial understanding? a systematic analysis from data to architecture.
\newblock \emph{arXiv preprint arXiv:2509.02359}, 2025{\natexlab{d}}.

\bibitem[Zhang et~al.(2025{\natexlab{e}})Zhang, Ding, Dong, He, Lin, Tang, Zang, Cao, Lin, and Wang]{zhang2025sec}
Zhixiong Zhang, Shuangrui Ding, Xiaoyi Dong, Songxin He, Jianfan Lin, Junsong Tang, Yuhang Zang, Yuhang Cao, Dahua Lin, and Jiaqi Wang.
\newblock Sec: Advancing complex video object segmentation via progressive concept construction.
\newblock \emph{arXiv preprint arXiv:2507.15852}, 2025{\natexlab{e}}.

\bibitem[Zhang et~al.(2025{\natexlab{f}})Zhang, Zhu, Ge, Zhao, Zhou, Li, Feng, Yao, and Han]{zhang2025corewardingstableselfsupervisedrl}
Zizhuo Zhang, Jianing Zhu, Xinmu Ge, Zihua Zhao, Zhanke Zhou, Xuan Li, Xiao Feng, Jiangchao Yao, and Bo Han.
\newblock Co-reward: Self-supervised reinforcement learning for large language model reasoning via contrastive agreement.
\newblock \emph{arXiv preprint arXiv:2508.00410}, 2025{\natexlab{f}}.

\end{thebibliography}
}

\appendix
\clearpage
\setcounter{page}{1}
\section{Method Details}
\label{sec:app-81k}
\subsection{Statistics of \methodname-81k}
\methodname-81k consists of 81,053 samples in total. All samples belong to the self-supervised tasks in \cref{section:self-supervised-tasks}. The examples of all tasks are shown in \cref{fig:app-shuf}, \cref{fig:app-flip}, \cref{fig:app-crop}, \cref{fig:app-depth}, and \cref{fig:app-crop}. Considering that \textit{Shuffled Patch Reordering} and \textit{Flipped Patch Recognition} share similar image layouts and structures, we regard them as one large task and evenly mix them with three other tasks, each containing approximately 20k samples. It also benefits the even distribution of the two task categories: \textit{Depth-free} and \textit{Depth-Based}. The number of samples for each task is demonstrated in \cref{tab:app-ss}. The formulation details of each task are provided in \cref{sec:app-shuf}, \cref{sec:app-flip}, \cref{sec:app-crop}, \cref{sec:app-dep}, \cref{sec:app-pos}. 

\begin{table}[h]
\caption{The size of all five tasks in the RL training dataset.}
  \centering
  \setlength{\tabcolsep}{4pt}
  \scalebox{1}{
  \begin{tabular}{@{}c|c|c@{}}
    \toprule
    Category&  Task & Size \\
    \midrule
    \multirow{3}{*}{Depth-free} &
    \textit{Shuffled Patch Reordering} & 16,028 \\
   & \textit{Flipped Patch Recognition} & 4,005 \\
   & \textit{Cropped Patch Inpainting} & 20,200 \\
   \cmidrule{1-3}
   \multirow{2}{*}{Depth-based} &\textit{Regional Depth Ordering}& 20,620 \\
    &\textit{Relative Position Prediction} & 20,200 \\
    \bottomrule
  \end{tabular}
}
  \label{tab:app-ss}
\end{table}

\subsection{Shuffled Patch Reordering}
\label{sec:app-shuf}
\begin{figure}[h]
  \centering
  \includegraphics[width=\linewidth]{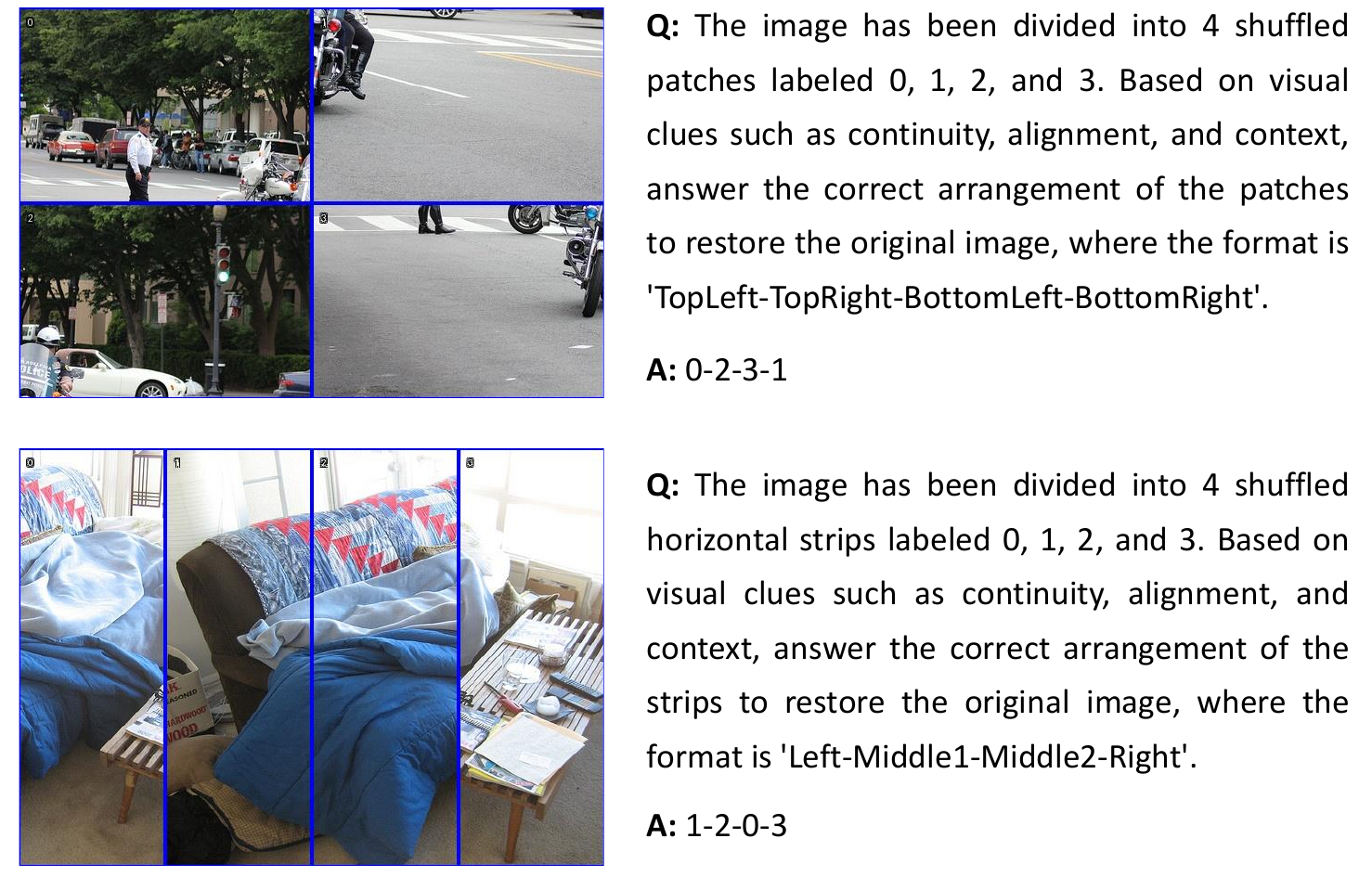}
   \caption{Examples of the task \textit{Shuffled Patch Reordering.}}
   \label{fig:app-shuf}
\end{figure}
We split the images into $M\times N$ patches, where $M$ is the number of patches in the vertical direction and $N$ is the number of patches in the horizontal direction. The examples are provided in \cref{fig:app-shuf}. For instance, the example on the bottom is patchified with $M=1, N=4$.

Besides, we may also apply a mask to one patch on some task samples. The details such as $M, N$, mask, etc., and the corresponding number of each type of samples are listed in \cref{tab:app-shuf}.

\begin{table}[h]
\caption{The details of \textit{Shuffled Patch Reordering} samples.}
  \centering
  \setlength{\tabcolsep}{4pt}
  \scalebox{1}{
  \begin{tabular}{@{}c|c|c@{}}
    \toprule
    Patchify Strategy&  Mask & Size \\
    \midrule
   $M$ = 2, $N$ = 2& \XSolidBrush
     & 4,000 \\
    \cmidrule{1-3}
  $M$ = 2, $N$ = 2 & \Checkmark & 4,028 \\
  \cmidrule{1-3}
   Horizontal ($M$ = 1, $N$ = 3 or 4)  &  \XSolidBrush& 4,991\\
 \cmidrule{1-3}
 Vertical ($M$= 3 or 4, $N$ = 1) &  \XSolidBrush& 3,009 \\
    \bottomrule
  \end{tabular}
}
  \label{tab:app-shuf}
\end{table}

\subsection{Flipped Patch Recognition}
\label{sec:app-flip}
\begin{figure}[h]
  \centering
  \includegraphics[width=\linewidth]{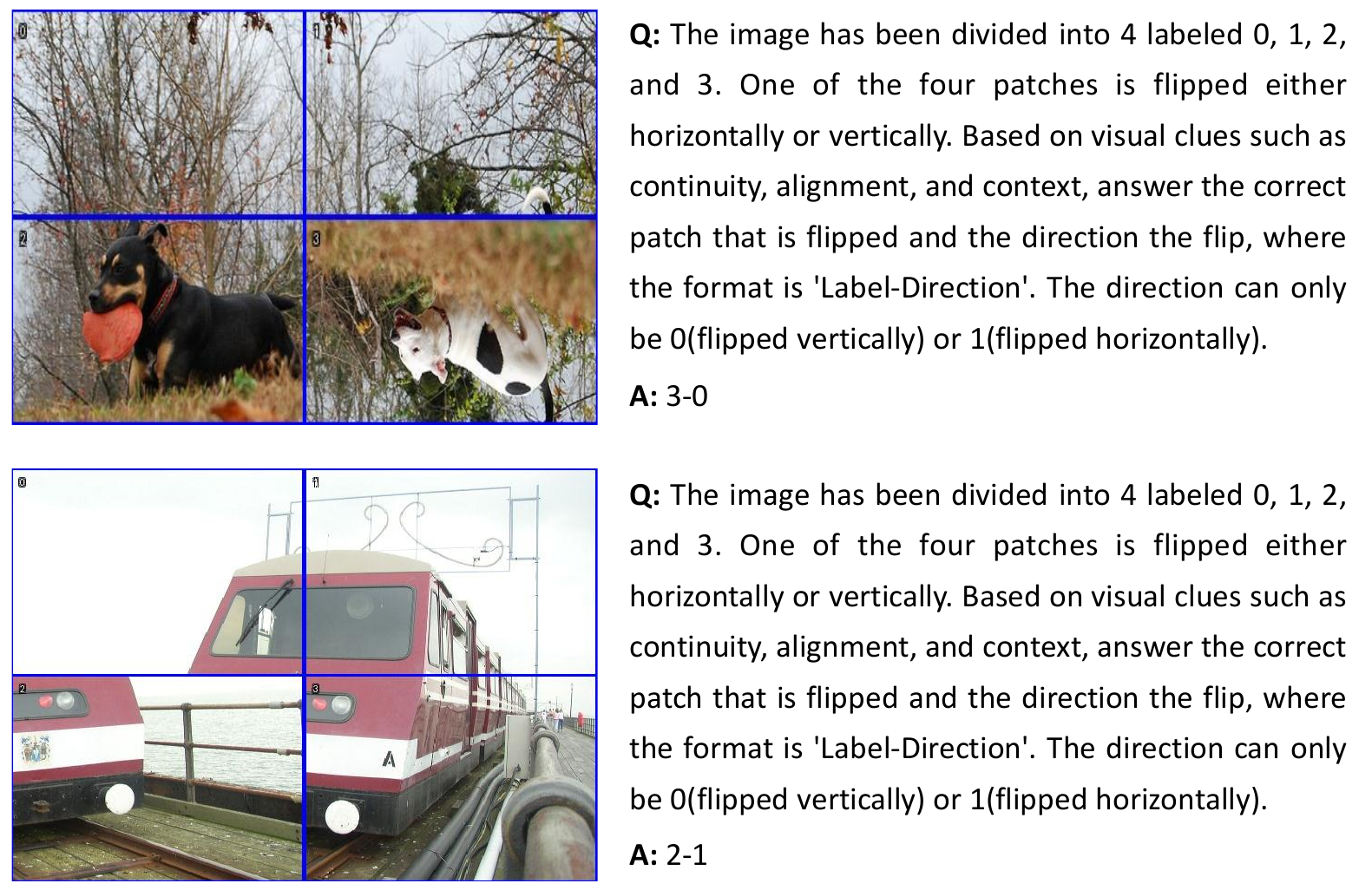}
   \caption{Examples of the task \textit{Flipped Patch Recognition.}}
   \label{fig:app-flip}
\end{figure}
The examples of this task are provided in \cref{fig:app-flip}.

\subsection{Cropped Patch Inpainting}
\label{sec:app-crop}
\begin{figure}[h]
  \centering
  \includegraphics[width=\linewidth]{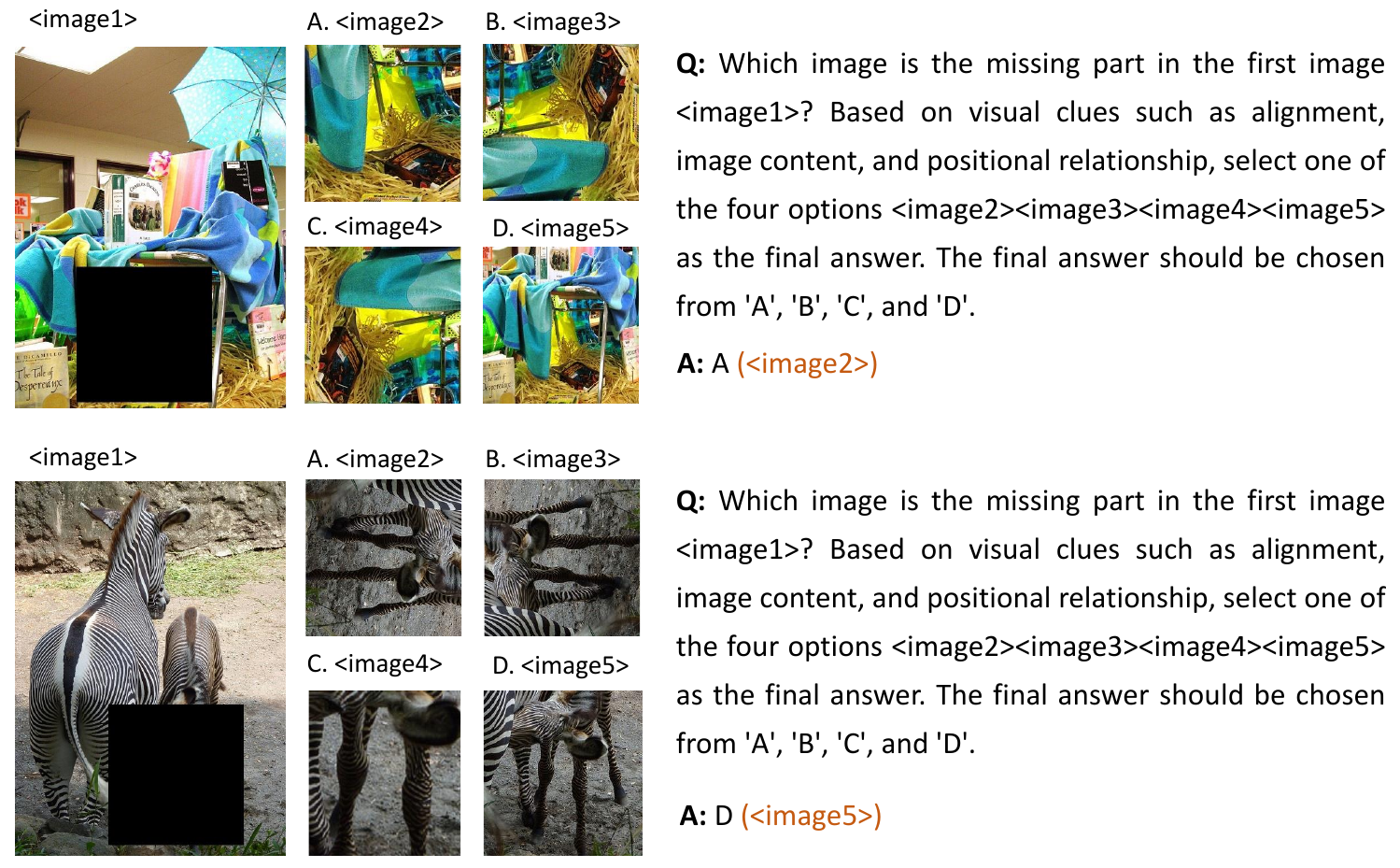}
   \caption{Examples of the task \textit{Cropped Patch Inpainting.}}
   \label{fig:app-crop}
\end{figure}
The examples are provided in \cref{fig:app-crop}. Among the four options in a problem, the correct option is the patch directly cropped from the blackened area of \texttt{<image1>}, retaining its original size. The other three options also come from \texttt{<image1>} to make the problem more challenging, and they are reshaped to the same size as the ground-truth patch. 

As illustrated in \cref{section:depth-free}, the three distractors are constructed in the following three forms: internal regions of the ground-truth, external regions of the ground-truth ($\theta$ = 0.25 or 0.5), and rotations of the ground-truth ($90^\circ$ clockwise or counterclockwise). We ensure that all three distractors are distinct from each other and the possibility of employing each method in formulating one problem is shown in \cref{tab:app-crop}.

\begin{table}[h]
\caption{The probability of adopting each method for constructing a distractor of one QA sample in \textit{Cropped Patch Inpainting} task.}
  \centering
  \setlength{\tabcolsep}{4pt}
  \scalebox{1}{
  \begin{tabular}{@{}c|c|c@{}}
    \toprule
    Method &  Parameter Value & Probability \\
    \midrule
    Internal Region & N/A & 0.2 \\
    \cmidrule{1-3}
    \multirow{2}{*}{External Region} &
    $\theta=0.25$ & 0.2 \\
   & $\theta=0.5$ & 0.2 \\
   \cmidrule{1-3}
   \multirow{2}{*}{Rotation} &$90^\circ$ Clockwise& 0.2 \\
    &$90^\circ$ Counterclockwise & 0.2 \\
    \bottomrule
  \end{tabular}
}
  \label{tab:app-crop}
\end{table}

\subsection{Regional Depth Ordering}
\label{sec:app-dep}

\begin{figure}[h]
  \centering
  \includegraphics[width=\linewidth]{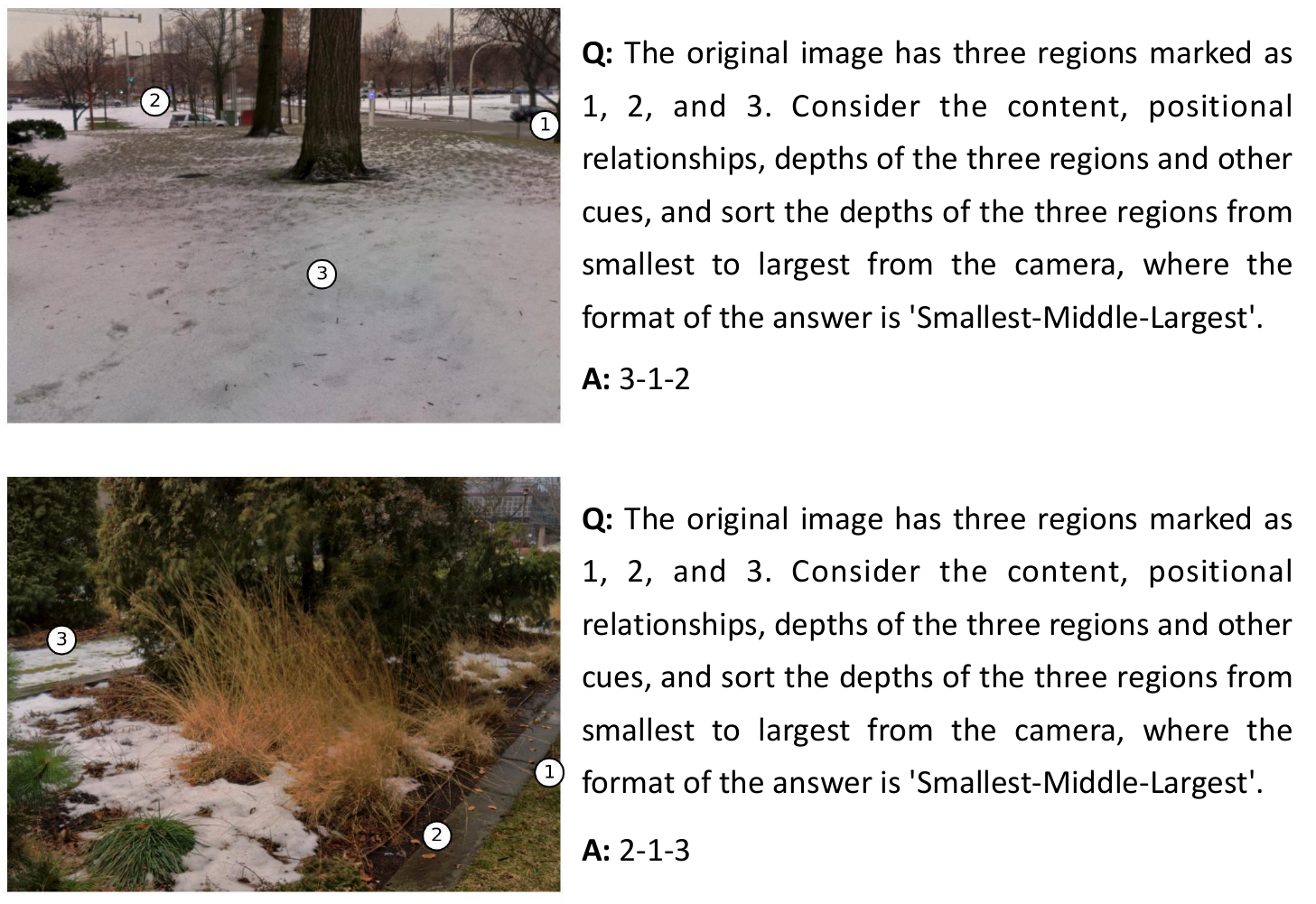}
   \caption{Examples of the task \textit{Regional Depth Ordering.}}
   \label{fig:app-depth}
\end{figure}
The examples of this task are provided in \cref{fig:app-depth}.

\subsection{Relative Position Prediction}
\label{sec:app-pos}

\begin{figure}[h]
  \centering
  \includegraphics[width=\linewidth]{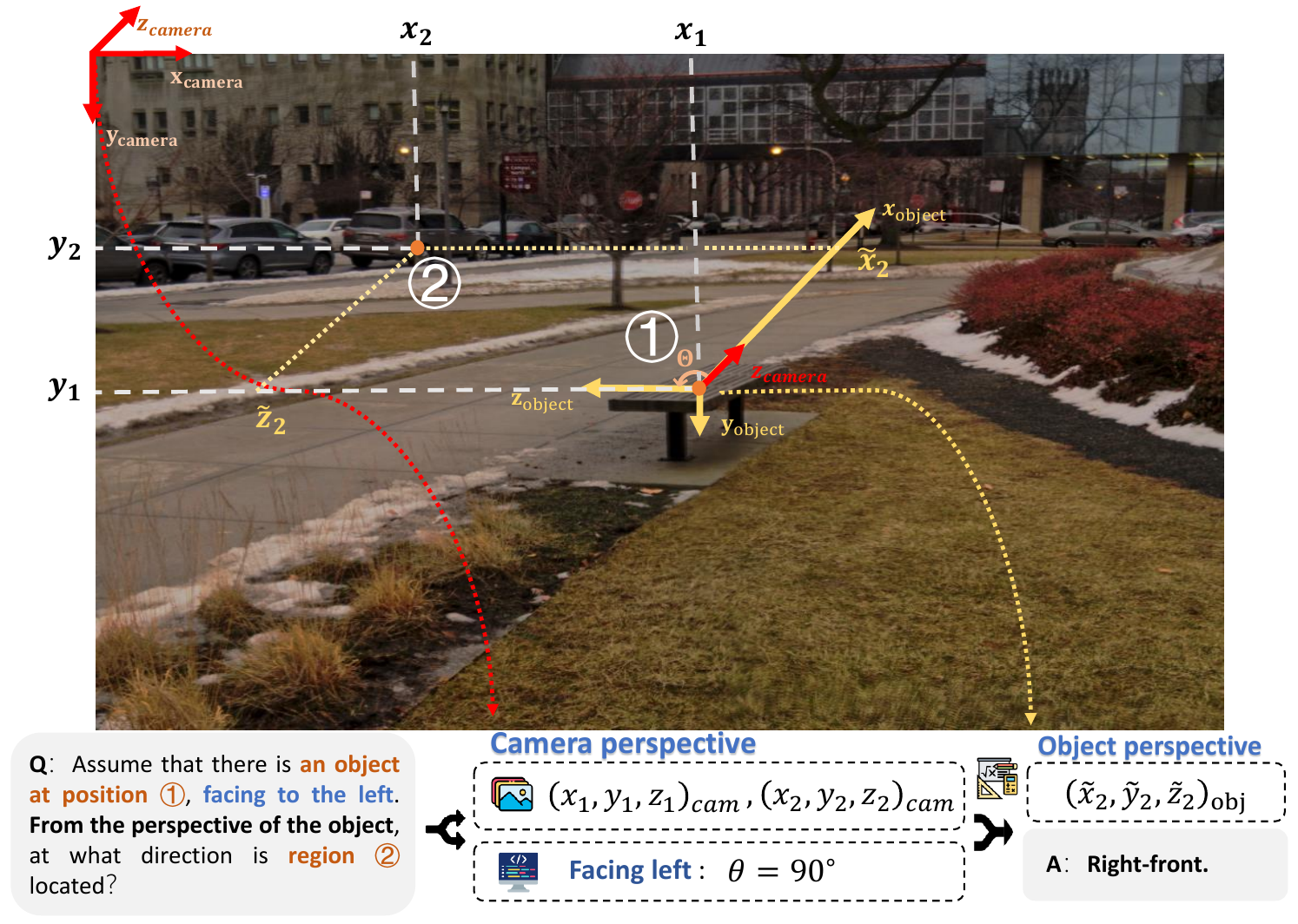}
   \caption{\textbf{The construction procedure of the task \textit{Relative position prediction}}. We define two coordinate systems based on the camera and the hypothesized object respectively. The $z$-axis represents the orientation. The $x$-axis represents the right side. The $y$-axis is always vertically downward. $(x_i,y_i,z_i)$ is the coordinate of position $i$ in the camera system while $(\tilde{x}_i, \tilde{y}_i, \tilde{z}_i)$ is defined in the object system.}
   \label{fig:orientation}
\end{figure}
\begin{figure}[h]
  \centering
  \includegraphics[width=\linewidth, height=5.9cm]{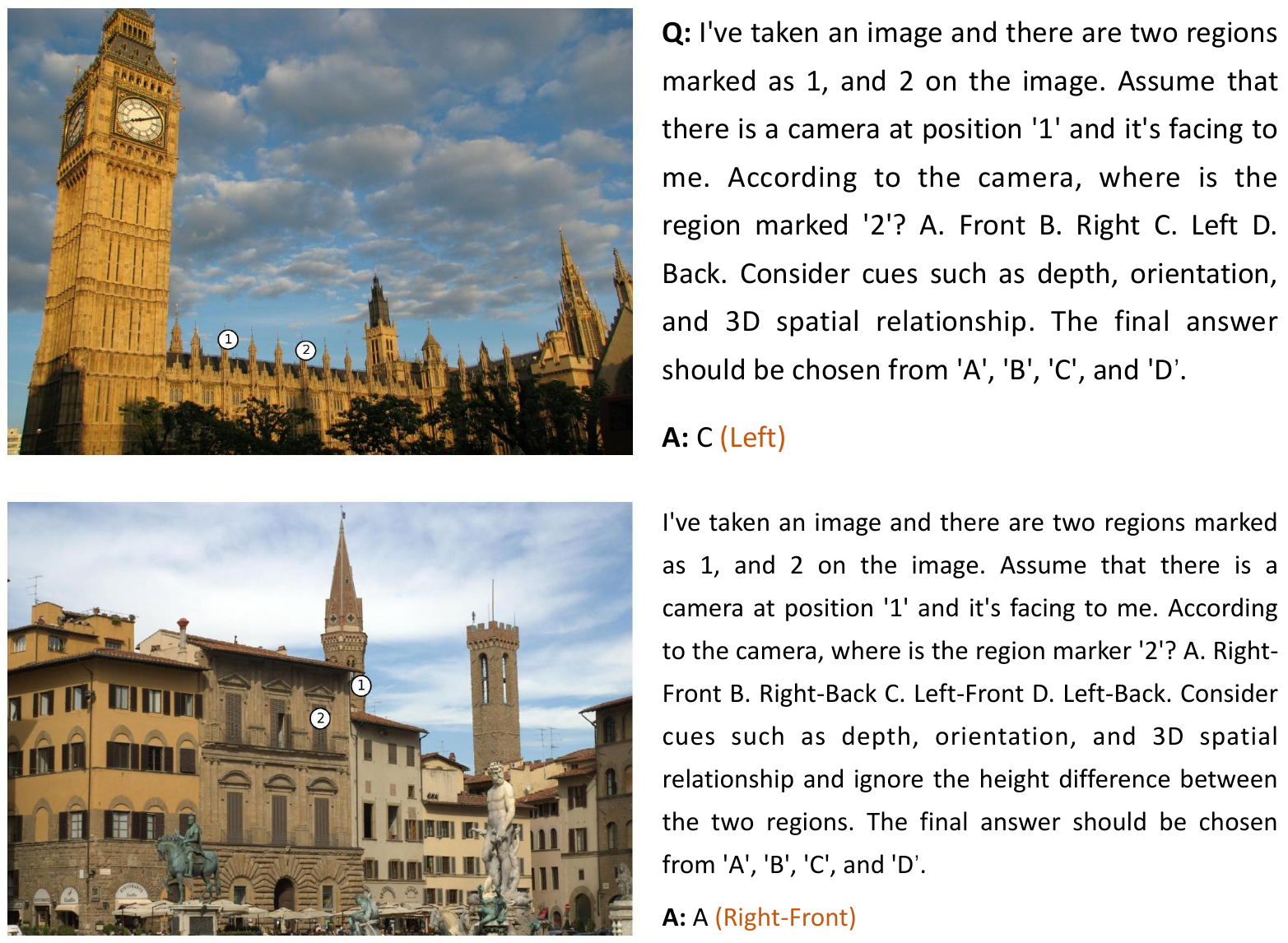}
   \caption{Examples of the task \textit{Relative Position Prediction}}
   \label{fig:app-pos}
\end{figure}

\Cref{fig:orientation} depicts the detailed construction procedure of this task, offering the definition of the variables used in \cref{section:depth-free} and explaining how the ground-truth answer can be derived through our automated pipeline. The examples are provided in \cref{fig:app-pos}.

\begin{table}[h]
\caption{The values of parameters in formulating \textit{Relative Position Prediction} samples. \textit{px} stands for pixels and \textit{nd} is the normalized depth in the raw RGB-D images, ranging from 0 to 1.}
  \centering
  \setlength{\tabcolsep}{4pt}
  \scalebox{1}{
  \begin{tabular}{@{}c|c|c| c@{}}
    \toprule
    Object Orientation &  $\theta$ & $\delta_x$ & $\delta_z$\\
    \midrule
   \multirow{2}{*}{$\vec{z}_{obj}\perp$ Image Plane}&$ 0^\circ$ & $150$ (px) 
     & $0.25$ (nd)\\
    \cmidrule{2-4}
  & $180^\circ$ & $150$ (px) 
     & $0.25$ (nd) \\
  \cmidrule{1-4}
   \multirow{2}{*}{$\vec{z}_{obj} \mathrel{/\mskip-2.5mu/}$ Image Plane} & $90^\circ$  &  $0.25$ (nd)
     & $150$ (px)\\
 \cmidrule{2-4}
 & $270^\circ$&  $0.25$ (nd)
     & $150$ (px) \\
    \bottomrule
  \end{tabular}
}
  \label{tab:app-pos}
\end{table}

During formulation, we define the parameters $\theta$ to represent the orientation of the object and $\delta_x,\delta_z$ as thresholds to avoid ambiguity of the ground-truth answers. $\delta_x$ is in the direction of $\vec{x}_{obj}$ and $\delta_z$ is in the direction of $\vec{z}_{obj}$. We use pixels as the unit for measuring in the direction parallel to the image plane (e.g., $\vec{x}_{cam}$), and normalized depth (0-1) as the unit for measuring in the direction perpendicular to the image plane (e.g., $\vec{z}_{cam}$). Since the orientation of the object changes due to the value of $\theta$, $\delta_x$ and $\delta_z$ are determined by their direction (parallel or perpendicular to the image plane). We set the thresholds as 150 pixels in the direction parallel to the image plane and 0.25 in the direction perpendicular to the image plane. The detailed values are shown in \cref{tab:app-pos}.

\subsection{Question Templates and Format Prompt}

We provide the question templates for the five self-supervised tasks. The words in \textcolor{ForestGreen}{green} are alternative content determined in the automated construction procedure.

\begin{tcolorbox}[colback=gray!10, colframe=black, colbacktitle=white!80!gray, coltitle=black, title= \textit{Shuffled Patch Reordering}, fonttitle=\bfseries]
\begin{enumerate}[label=\textbf{\arabic*.}]
    \item \textbf{M = 2, N = 2, w.o. mask}:
    
    \textbf{Question:} The image has been divided into 4 shuffled
patches labeled 0, 1, 2, and 3. Based on visual
clues such as continuity, alignment, and context,
answer the correct arrangement of the patches
to restore the original image, where the format is
`TopLeft-TopRight-BottomLeft-BottomRight'.\\

    \item \textbf{M = 2, N = 2, with mask:}

    \textbf{Question:} The image has been divided into 4 shuffled patches labeled 0, 1, 2, and 3. One of the four patches is masked completely by white pixels. Based on visual clues such as continuity, alignment, and context, answer the correct arrangement of the patches to restore the original image, where the format is `TopLeft-TopRight-BottomLeft-BottomRight'.\\
    \item \textbf{M = 1, N = 3 or 4, w.o. mask:} 
    
    \textbf{Question:} The image has been divided into 3\textcolor{ForestGreen}{(4)} shuffled horizontal strips labeled 0, 1, 2, and 3. Based on visual clues such as continuity, alignment, and context, answer the correct arrangement of the strips to restore the original image, where the format is `Left-Middle-Right'\textcolor{ForestGreen}{(`Left-Middle1-Middle2-Right')}.\\
    
    \item \textbf{M = 3 or 4, N = 1, w.o. mask:}
    
    \textbf{Question:} The image has been divided into 3\textcolor{ForestGreen}{(4)} shuffled vertical strips labeled 0, 1, 2, and 3. Based on visual clues such as continuity, alignment, and context, answer the correct arrangement of the strips to restore the original image, where the format is `Top-Middle-Bottom'\textcolor{ForestGreen}{(`Top-Middle1-Middle2-Bottom')}.
\end{enumerate}

\end{tcolorbox}

\begin{tcolorbox}[colback=gray!10, colframe=black, colbacktitle=white!80!gray, coltitle=black, title= \textit{Flipped Patch Recognition}, fonttitle=\bfseries]
\textbf{Question:} The image has been divided into 4 labeled 0, 1, 2, and 3. One of the four patches is flipped either horizontally or vertically. Based on visual clues such as continuity, alignment, and context, answer the correct patch that is flipped and the direction the flip, where the format is 'Label-Direction'. The direction can only be 0(flipped vertically) or 1(flipped horizontally). 

\end{tcolorbox}

\begin{tcolorbox}[colback=gray!10, colframe=black, colbacktitle=white!80!gray, coltitle=black, title= \textit{Cropped Patch Inpainting}, fonttitle=\bfseries]
\textbf{Question:} Which image is the missing part in the first image $<$image1$>$? Based on visual clues such as alignment, image content, and positional relationship, select one of the four options $<$image2$>$$<$image3$>$$<$image4$>$$<$image5$>$ as the final answer. The final answer should be chosen from `A', `B', `C', and `D'.

\end{tcolorbox}

\begin{tcolorbox}[colback=gray!10, colframe=black, colbacktitle=white!80!gray, coltitle=black, title= \textit{Regional Depth Ordering}, fonttitle=\bfseries]
\textbf{Question:} The original image has three regions marked as 1, 2, and 3. Consider the content, positional relationships, depths of the three regions and other cues, and sort the depths of the three regions from smallest to largest from the camera, where the format of the answer is `Smallest-Middle-Largest'.

\end{tcolorbox}

\begin{tcolorbox}[colback=gray!10, colframe=black, colbacktitle=white!80!gray, coltitle=black, title= \textit{Regional Depth Ordering}, fonttitle=\bfseries]
\textbf{Question:} I've taken an image and there are two regions marked as 1, and 2 on the image. Assume that there is a camera at position `1'\textcolor{ForestGreen}{(`2')} and it's facing to the left of the image. According to the camera, where is the region marked `2'\textcolor{ForestGreen}{(`1')}? A. Front \textcolor{ForestGreen}{(Right-Front)} B. Right\textcolor{ForestGreen}{(Right-Back)} C. Back\textcolor{ForestGreen}{(Left-Front)} D. Left\textcolor{ForestGreen}{(Left-Back)}. Consider cues such as depth, orientation, and 3D spatial relationship. The final answer should be chosen from `A', `B', `C', and `D'.

\end{tcolorbox}

During GRPO training, a format prompt is appended at the end of the task questions to enable the model to generate reasoning content. By following a fixed format, it's also easier to extract the final answers from the model's responses for computing the accuracy reward.
\begin{tcolorbox}[colback=gray!10, colframe=black, colbacktitle=white!80!gray, coltitle=black, title= Format Prompt for Training, fonttitle=\bfseries]
You FIRST think about the reasoning process as an internal monologue and then provide the final answer. The reasoning process MUST BE enclosed within $\langle$think$\rangle$ $\langle$/think$\rangle$ tags. The final answer MUST BE put in \verb|\|boxed\{\}.

\end{tcolorbox}
\section{Evaluation Details}
\label{sec:app-exp}

\begin{table*}[t]
  \caption{Performance of Qwen3-VL-4B (baseline model) and \methodname-4B on spatial understanding.}
  \centering
  \vspace{-10pt}
  \setlength{\tabcolsep}{2pt}
  \scalebox{0.90}{
  \normalsize
  \begin{tabular}{@{}lccccccccc@{}}
    \toprule
    \multirow{2}{*}{Models} & \multirow{2}{*}{Reasoning} & \multicolumn{6}{c}{Image} & \multicolumn{1}{c}{Video}  & \multirow{2}{*}{Avg.}  \\
\cmidrule(lr){3-8} \cmidrule(l){9-9} 
  & & Spatial457 & 3DSRBench & SpatialEval & $\text{QSpatial}_{plus}$ &{What'sUp}& ViewSpatial  & {VSI-Bench}\\
    \midrule
    Qwen3-VL-4B & \XSolidBrush & 53.43 & 56.46 & 63.04 & 63.37 & 98.78 & 39.09  & 46.82 & 60.14  \\
    Qwen3-VL-4B & \Checkmark & 55.25 & 55.83 & 71.69 & 61.39 & 96.83 & 41.79 & 38.82 & 60.23\\
    \methodname-4B & \Checkmark & 57.12 & 59.48 & 72.38 & 59.41 & 97.44& 42.07 & 42.13 & \textbf{61.43} \textbf{\textcolor[RGB]{0, 120, 50}{(+1.29)}} \\
    \bottomrule
  \end{tabular}
  }
  \vspace{-12pt}
  \label{tab:app-v3-spa}
\end{table*}

\subsection{Spatial Benchmarks}
\textbf{Benchmarks supported in VLMEvalkit.} VLMEvalkit supports \textit{3DSRBench}, \textit{SpatialEval}, \textit{QSpatial-plus}, and \textit{Spatial457}. The former two benchmarks are multiple-choice problems with a non-CoT original prompt. Therefore, we use the official code in VLMEvalkit to evaluate the reasoning-free settings in \cref{tab:spa} and add our training format prompt to evaluate all reasoning-required settings.

\textbf{\textit{QSpatial-plus}} targets the quantitative prediction of 3D distances and requires a strict output format encompassing \textit{scalar} and \textit{distance unit} to facilitate its final score computation. Instead of employing our format prompt, we follow the official prompts (including both non-reasoning and reasoning version) during the evaluation of all models.

\textbf{\textit{Spatial457}} is also not in the form of multiple-choice questions, and its prompt requires CoT response. So we use the original prompt for reasoning-required baseline settings in \cref{tab:spa}, and our format prompt for \methodname-3B (and 7B) to ensure consistency with training. For reasoning-free baseline settings, the prompt to enable direct outputs of the answers is: \textit{Please directly give the answer.}

\noindent
\textbf{Other Benchmarks.} \textit{What'sUp, ViewSpatial, and VSI-Bench} are not supported in VLMEvalkit. Our evaluation implementation makes minor modification to the official code to adapt it to Qwen2.5-VL architecture while strictly preserving the original metrics and evaluation procedures.

\textbf{\textit{What'sUp}} contains multiple-choice problems targeting the recognition of unambiguous 2D spatial relation of two objects in an image (e.g., A mug under a table). The evaluation metric is exact matching of the option letter. We apply the following format prompt for evaluating non-reasoning settings in \cref{tab:spa}: \textit{Based on the image, choose the correct option from the list below.}, and append our training format prompt for testing reasoning settings.

\textbf{\textit{ViewSpatial}} aims at evaluating multi-perspective spatial reasoning, requiring the model's capability of 3D reconstruction and perspective transformation. Similarly, all problems are in the form of multiple-choice questions. We use the official code in our experiment. However, the original prompt doesn't explicitly instruct the model to generate a reasoning process, but it also doesn't guide the model to directly output the final answer. To accommodate it to our experiment settings, we define both the non-reasoning and reasoning format prompts as follows:
\begin{itemize}
    \item Prompt for Reasoning-free settings: \textit{Reply only to the corresponding option.}\verb|\|\textit{nAnswer:}
    \item Prompt for Reasoning settings: \textit{The final answer should be the option letter from the given choices.}\verb|\|\textit{n \textcolor{Mahogany}{ + Format Prompt for Training}}
\end{itemize}

\textbf{\textit{VSI-Bench}} targets spatial understanding of egocentric videos. It contains multiple-choice answers and numerical answers
 format. We follow the metrics proposed by the benchmark \cite{yang2025thinkingspacemultimodallarge}, which uses exact matching for multiple-choice answers and Mean Relative Accuracy ($MRA$) for numerical answers. Given a numerical model prediction $\hat{y}$ and its corresponding ground-truth value, $MRA$ is defined with a confidence threshold set $\mathcal{C}=\{0.05, 0.10,..., 0.5\}$: $MRA=\sum\limits_{x\in\mathcal{C}} \mathds{1} (\frac{|\hat{y}-y|}{y}<x)$.
 To balance efficiency and video quality, set \textit{max frames} to $128$ for each video input. 
 
 Both non-reasoning and reasoning format prompts for VSI-Bench are given as follows:
\begin{itemize}
    \item Prompt for Reasoning-free settings: \textit{Answer directly with a number(integer or decimal). / Answer directly with the option letter from the given choices.}
    \item Prompt for Reasoning settings: \textit{The final answer should be a number(integer or decimal).}\verb|\|\textit{n} \textcolor{Mahogany}{ \textit{+ Format Prompt for Training}} / \textit{The final answer should be the option letter from the given choices.}\verb|\|\textit{n} 
    \textcolor{Mahogany}{\textit{ + Format Prompt for Training}}
\end{itemize}

\subsection{General Visual Benchmarks}
\textbf{Evaluation Implementation.} All the benchmarks in \cref{sec:exp-gen} for testing models' general visual capabilities are supported in VLMEValkit. And we use it to implement the entire evaluation of all models on these benchmarks. 

\noindent
\textbf{Baseline Models.} The baseline models are evaluated by directly applying the original prompt provided in the toolkit. 

\noindent
\textbf{Our Models.} For our models (\methodname-3B and \methodname-7B), we employ the original prompt in general VQA benchmarks as we discover that the reasoning process hardly yields benefits in such problems only requiring simple visual perception, and we append the training format prompt for \textit{OCR and chart understanding} benchmarks for consistency with training since they demand some basic analysis (e.g., numeric comparison, calculation) as well as fine-grained comprehension of rich visual details, which shares similarities to our depth-free tasks.

\subsection{Results on Qwen3-VL-based Models}
\label{sec:app-qwen3}
We train \methodname-4B, initialized from Qwen3-VL-4B-Instruct, on our dataset \methodname-81k solely composed of self-supervised QA samples. To make the evaluation consistent with \cref{tab:spa} and \cref{tab:general} in \cref{sec:exp}, we evaluate both the non-reasoning and reasoning variants of the baseline model for spatial understanding benchmarks and compare them with \methodname-4B. The results are shown in \cref{tab:app-v3-spa}. The results of baseline models and our model on general VQA benchmarks are provided in \cref{tab:v3-general}.

\begin{table}[h]
\caption{Performance of Qwen3-VL-4B (baseline model) and \methodname-4B on general VQA. Our model has achieved an average accuracy gain of 1.18\%.}
  \centering
  \setlength{\tabcolsep}{3pt}
  \scalebox{0.97}{
  \begin{tabular}{@{}l|c|c|c|c|c@{}}
    \toprule
    Models &  MMBench & BLINK& Hallusion & RealWorld & Avg. \\
    \midrule
    Qwen3 & 84.10 & 64.81 & 56.38 & 71.11 & 69.10 \\
    Ours & 84.26 & 65.23 & 62.23 & 69.41 & 70.28 \\
    \bottomrule
  \end{tabular}
}
  \label{tab:v3-general}
\end{table}

\section{Qualitative Analysis}
\label{sec:app-qualitative}
In this section, we provide some qualitative examples to compare the performance of  \methodname-7B and our base model Qwen2.5-VL-7B on spatial understanding problems. These examples cover a broad range of spatial intelligence: 3D location understanding, depth comparison, orientation recognition, multi-object relationship perception, and reasoning on viewpoint transformation.
\begin{figure*}[h]
  \centering
  \includegraphics[width=0.7\linewidth]{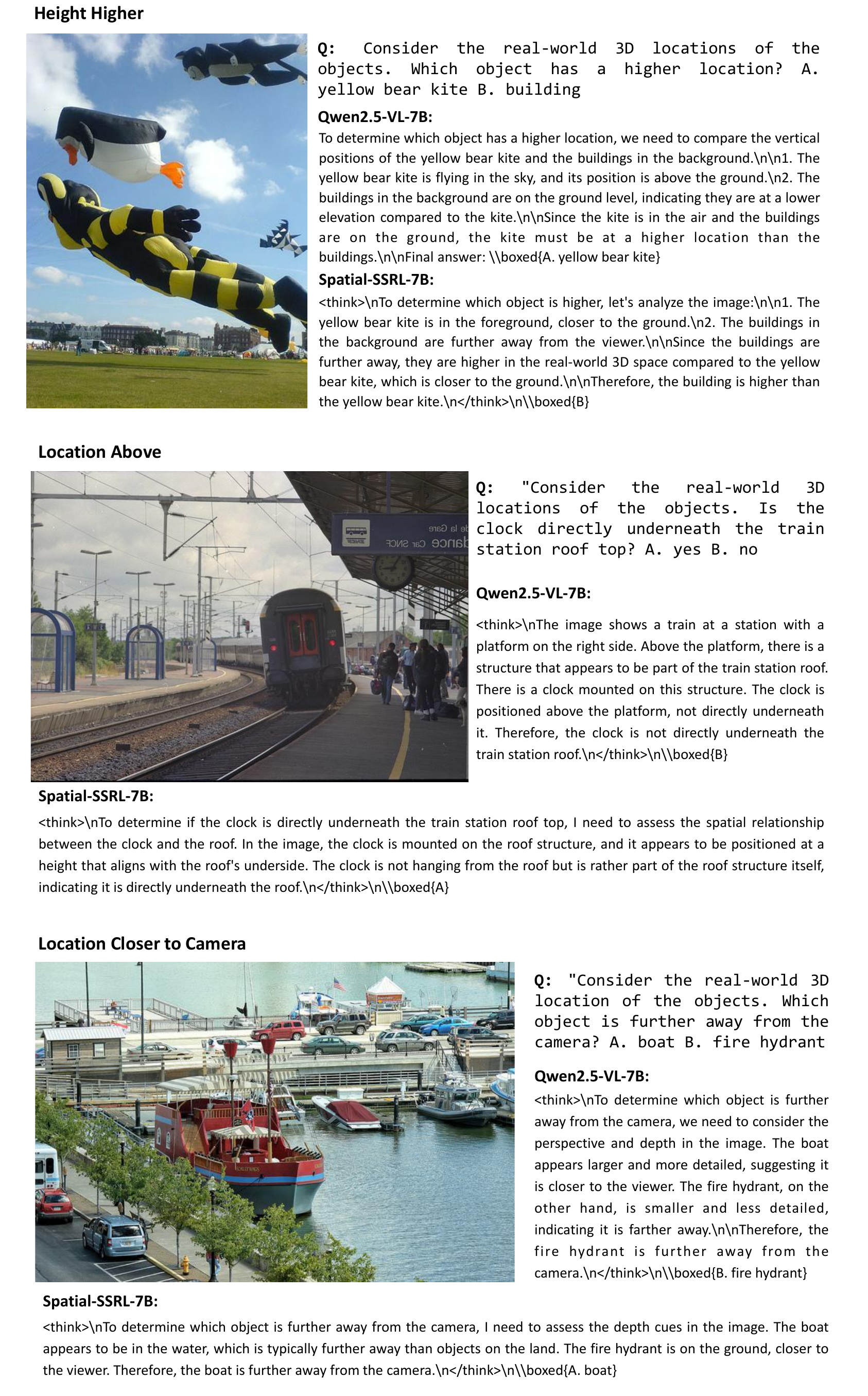}
   \caption{Qualitative examples of spatial understanding}
   \label{fig:long}
\end{figure*}
\begin{figure*}[t]
  \centering
  \includegraphics[width=0.7\linewidth]{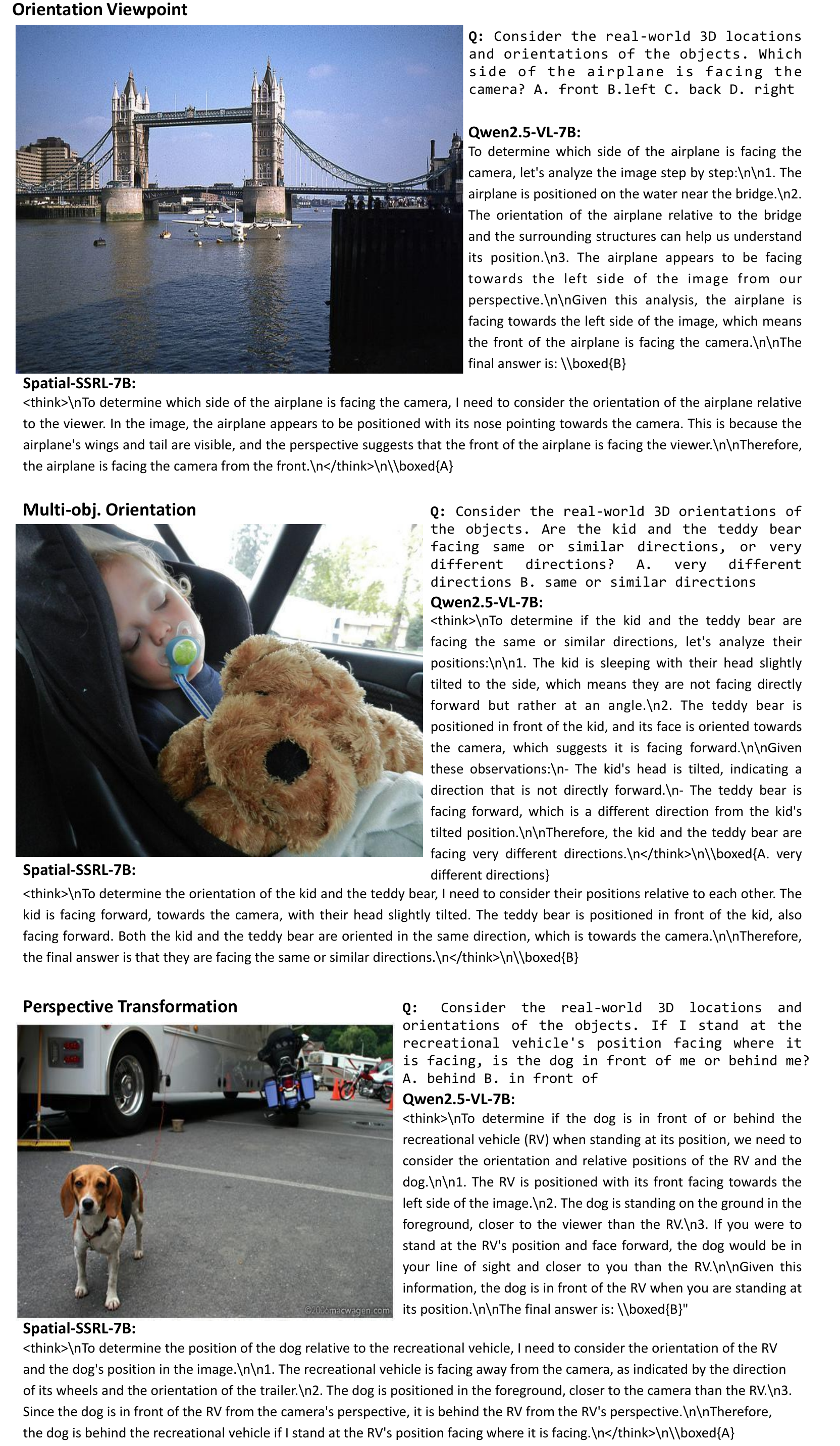}
   \caption{Qualitative examples of spatial understanding}
   \label{fig:long_2}
\end{figure*}

\end{document}